\documentclass{article}

\usepackage{microtype}
\usepackage{graphicx}
\usepackage{subcaption}
\usepackage{enumitem}
\usepackage{booktabs} 
\usepackage{xcolor,colortbl,booktabs,tikz,tcolorbox,adjustbox,enumitem}
\usepackage{times}

\definecolor{clBinClaude}{HTML}{E65100}
\definecolor{clGreen}{HTML}{2E7D32}
\definecolor{clRed}{HTML}{C62828}
\definecolor{clLightGray}{HTML}{FAFAFA}
\definecolor{clHuman}{HTML}{37474F}
\definecolor{clBinGPT}{HTML}{FF8A65}
\definecolor{clGEvalGPT4}{HTML}{283593}
\definecolor{clGEvalGPT}{HTML}{7986CB}
\definecolor{clUniEvalT5}{HTML}{00695C}
\definecolor{clUniEvalGPT}{HTML}{80CBC4}

\tcbset{
  casebox/.style={
    colback=clLightGray,
    colframe=black!40,
    fonttitle=\bfseries\small,
    boxrule=0.4pt,
    arc=2pt,
    left=4pt, right=4pt, top=3pt, bottom=3pt,
  }
}
\newcommand{\yes}{\textcolor{clGreen}{\textbf{Y}}}
\newcommand{\no}{\textcolor{clRed}{\textbf{N}}}

\usepackage[accepted]{icml2026}

\usepackage{amsmath}
\usepackage{amssymb}
\usepackage{mathtools}
\usepackage{amsthm}

\expandafter\let\csname algorithm*\endcsname\relax
\expandafter\let\csname endalgorithm*\endcsname\relax

\usepackage[ruled,vlined,linesnumbered]{algorithm2e}
\SetAlgoNlRelativeSize{0}

\usepackage[capitalize,noabbrev]{cleveref}
\usepackage[section]{placeins}
\theoremstyle{plain}

\theoremstyle{definition}

\theoremstyle{remark}

\usepackage[textsize=tiny]{todonotes}
\usepackage{xspace}

\newcommand{\bindeval}{\textsc{BinEval}\xspace}
\newcommand{\dashscore}{\textemdash}
\newcommand{\rwparagraph}[1]{\vspace{0.2ex}\noindent\textbf{#1.} }

\icmltitlerunning{Ask, Don’t Judge: Binary Questions for Interpretable LLM Evaluation and Self-Improvement}

\begin{document}

\twocolumn[
 
    \icmltitle{Ask, Don’t Judge: Binary Questions for Interpretable LLM Evaluation and Self-Improvement}
  \begin{icmlauthorlist}
    \icmlauthor{Sangwoo Cho}{capone}
    \icmlauthor{Kushal Chawla}{capone}
    \icmlauthor{Pengshan Cai}{capone}
    \icmlauthor{Zefang Liu}{capone}
    \icmlauthor{Chenyang Zhu}{capone}
    \icmlauthor{Shi-Xiong Zhang}{capone}
    \icmlauthor{Sambit Sahu}{capone}
  \end{icmlauthorlist}

  \icmlaffiliation{capone}{Capital One, AI Foundations, McLean, VA 22102, USA}
\icmlcorrespondingauthor{Sangwoo Cho}{sangwoo.cho@capitalone.com}
  \icmlkeywords{large language models, evaluation, prompt optimization, interpretability}

  \vskip 0.3in
]
\printAffiliationsAndNotice{\hspace{0pt}}

\begin{abstract}
Evaluating LLM outputs remains a major bottleneck in NLP: human evaluation is expensive and slow, lexical metrics correlate poorly with human judgments on open-ended generation, and holistic LLM judges often produce opaque scores that are hard to debug. We propose \bindeval, a framework that decomposes evaluation criteria into atomic binary questions and aggregates the resulting verdicts into interpretable, multi-dimensional scores. Given a task prompt, a meta-prompt generates fine-grained evaluation questions, and an LLM answers them independently for each output, yielding transparent question-level feedback together with calibrated overall scores. This decomposition makes evaluation easier to inspect, easier to diagnose, and directly usable for prompt improvement.
Across SummEval, Topical-Chat, and QAGS, \bindeval matches or outperforms strong baselines including UniEval and G-Eval, with especially strong results on factual consistency benchmarks such as QAGS. Beyond competitive correlation with human judgments, \bindeval better matches human score distributions and avoids the ceiling effects common in prior LLM judges, leading to better discrimination between borderline and clearly flawed outputs. We further show that the same question-level feedback supports iterative prompt optimization, improving evaluator prompts on summarization and generation prompts on IFBench under both self-update and cross-model update settings. Overall, \bindeval provides a task-agnostic, training-free, and interpretable evaluation framework that combines strong empirical performance with practical diagnostic and optimization value.
\end{abstract}

\section{Introduction}

The rapid progress of large language models (LLMs) has made generation easy and evaluation hard. Modern systems can produce fluent, contextually appropriate outputs across tasks such as summarization, dialogue, reasoning, and instruction following, but evaluating those outputs remains a major bottleneck. Human evaluation is slow and expensive, lexical metrics such as ROUGE~\cite{lin04}, BLEU~\cite{papineni02}, and BERTScore~\cite{zhang20} miss semantic correctness and factuality, and holistic LLM judges~\cite{zheng23,liu23} often return opaque scores that are difficult to diagnose.

This bottleneck is especially costly in iterative development. Comparing prompts, models, or decoding strategies requires feedback that is not only accurate but also actionable. A single scalar score is often insufficient: if a summary receives a mediocre rating, it is still unclear whether the problem is factual inconsistency, weak relevance, missing content, or poor fluency.

Our premise is simple: instead of asking a model for one broad judgment, ask it a set of small, checkable questions. We therefore propose \bindeval, which decomposes each evaluation criterion into atomic yes/no questions and aggregates the resulting verdicts into interpretable scores. This decomposition turns evaluation from a black-box verdict into a structured diagnostic signal, making it easier to inspect, debug, and improve both evaluators and generators.

\bindeval has three components. First, a meta-prompt decomposes a task prompt into atomic questions organized by evaluation dimension. Second, an evaluator answers each question independently and aggregates the answers into per-dimension and overall scores. Third, a two-phase optimization loop improves both evaluator prompts and generation prompts using question-level feedback.

We evaluate \bindeval on SummEval~\cite{fabbri21}, Topical-Chat~\cite{mehri20}, and QAGS~\cite{wang20qags}, and we study iterative prompt updating on summarization and IFBench.

Our contributions are:
\begin{itemize}[itemsep=1pt, topsep=1pt]
    \item \textbf{A general framework for interpretable evaluation.} We decompose evaluation criteria into atomic yes/no questions, yielding a task-agnostic and modular method.
    \item \textbf{Strong performance without task-specific training.} \bindeval matches or exceeds trained evaluators and holistic LLM judges on SummEval, Topical-Chat, and QAGS.
    \item \textbf{Iterative prompt improvement.} We introduce a two-phase optimization loop that improves prompts for both summarization and IFBench.
    \item \textbf{Debuggable scores.} Each \bindeval score is grounded in individual verdicts with explanations, making evaluator behavior easier to inspect and diagnose.
\end{itemize}

\section{Related Work}

\rwparagraph{Traditional Evaluation Metrics}
Lexical overlap metrics--ROUGE~\cite{lin04}, BLEU~\cite{papineni02}, and METEOR~\cite{banerjee05}--remain standard for summarization and translation evaluation, but they often struggle to capture semantic equivalence in open-ended generation. Embedding-based metrics such as BERTScore~\cite{zhang20} and MoverScore~\cite{zhao19} improve semantic matching by operating in representation space, while generation-based metrics like BARTScore~\cite{yuan21} frame evaluation as text generation. More recent reference-free methods like ParaPLUIE~\cite{lemesle25parapluie} measure meaning preservation using model perplexity without requiring gold references, and frameworks like OmniScore~\cite{alam26omniscore} use deterministic learned evaluators to support scalable multilingual assessment.

\rwparagraph{LLM-as-Judge}
Recent work has increasingly leveraged LLMs themselves as evaluators. G-Eval~\cite{liu23} uses chain-of-thought reasoning followed by a Likert-scale rating, while AlpacaEval~\cite{li23} and MT-Bench / Chatbot Arena~\cite{zheng23} rely on pairwise or preference-based judgments. The paradigm has also expanded to specialized open-source evaluators such as Prometheus 2~\cite{kim24prometheus2}, which approximates the depth of human and proprietary model judgments. However, these judges remain susceptible to position, verbosity, and self-enhancement biases~\cite{zheng23}. Recent benchmarks like JudgeBiasBench~\cite{zhou26judgebiasbench} further systematize these concerns by providing a taxonomy of judge biases and proposing debiasing strategies.

\rwparagraph{Multi-Dimensional Evaluation}
Multi-dimensional evaluation aims to decompose quality into interpretable facets such as coherence, faithfulness, informativeness, and relevance. UniEval~\cite{zhong22} is a key prior example: it reformulates evaluation as Boolean question answering and fine-tunes a T5-based evaluator for multiple dimensions. More recent work similarly decomposes evaluation into facets like informativeness and faithfulness~\cite{alam26omniscore}, while hybrid frameworks such as QAEval~\cite{yue25qaeval} combine rule-based reliability with a Mixture of Evaluators for open-ended generation tasks. Together, these methods reinforce the value of breaking evaluation into smaller, more structured judgments.

\rwparagraph{Atomic Decomposition for Evaluation}
FActScore~\cite{min23} pioneered the ``decompose-then-verify'' paradigm by breaking long-form generations into atomic facts and verifying them individually. Related frameworks such as ARES~\cite{saadfalcon24} and RAGAS~\cite{es24} extend similar decomposition ideas to retrieval-augmented generation, while OpenFActScore~\cite{lage25openfactscore} enables open-source fact-checking with atomic evaluation. These approaches demonstrate that fine-grained decomposition can improve factual assessment, although they typically decompose generated content rather than evaluation criteria themselves.

\rwparagraph{Prompt Optimization}
Prompt optimization has increasingly shifted from manual instruction engineering toward automated and programmatic refinement. DSPy~\cite{khattab23} provides a framework for declarative, self-improving language-model pipelines, and algorithms like MIPRO~\cite{opsahlong24mipro} perform Bayesian search over instructions and demonstrations. OPRO~\cite{yang23} and APE~\cite{zhou23ape} likewise use language models to iteratively generate and refine prompts. More recent methods such as MARS~\cite{zhang25mars} introduce multi-agent Socratic optimization, while LLM-AutoDiff~\cite{yin25llmautodiff} treats textual inputs as trainable parameters in graph-structured workflows. These methods motivate our use of disagreement-driven prompt refinement as a targeted optimization signal.

\section{Method}

We present \bindeval in three parts: binary question generation (\cref{sec:qgen}), binary evaluation and scoring (\cref{sec:beval}), and iterative prompt optimization (\cref{sec:evalopt,sec:genopt}).

\subsection{Binary Question Generation}
\label{sec:qgen}

Let $T$ denote a task prompt defining the generation requirements, such as a summarization instruction, a dialogue system prompt, or an instruction-following specification. We define a \emph{decomposition function} that maps $T$ to a set of binary questions:
\[
\mathcal{Q} = \mathcal{F}_{\text{LLM}}(T; M) = \{q_1, q_2, \dots, q_N\}.
\]
where $M$ is a meta-prompt that instructs an LLM to perform a two-step decomposition.

\rwparagraph{Step 1 -- Summarize}
We first summarize the task prompt $T$ into an explicit set of requirements $\mathcal{R} = \{r_1, r_2, \dots, r_K\}$. Each requirement $r_k$ captures a distinct evaluation criterion, such as whether the output includes a key piece of information or obeys a formatting constraint. This summarization step is intended to help the model form a coherent representation of the full task before attempting finer-grained decomposition.

\rwparagraph{Step 2 -- Decompose}
For each requirement $r_k$, we generate one or more binary questions such that answering ``yes'' indicates the output satisfies the requirement and answering ``no'' indicates a violation. Requirements that implicitly contain multiple sub-tasks are decomposed into separate questions, and each question is paired with a concise violation example to clarify the negative case. This design is motivated by prior work showing that complex reasoning is often improved by decomposing a task into simpler sub-problems that can be solved sequentially or modularly~\cite{zhou22leasttomost,khot22decomposed}. In our setting, the same intuition suggests that evaluation becomes easier when the model answers targeted binary questions about simplified sub-tasks rather than making a single holistic judgment.


The questions can be organized into evaluation dimensions. For a set of dimensions $\mathcal{D}$, such as coherence, consistency, fluency, and relevance, the questions partition as
\[
\mathcal{Q} = \bigcup_{d \in \mathcal{D}} \mathcal{Q}_d,
\]
where $\mathcal{Q}_d$ contains questions specific to dimension $d$. The meta-prompt $M$ is task-agnostic: the same meta-prompt generates appropriate binary questions for summarization, dialogue, instruction following, or any other task, with only $T$ changing.

\subsection{Binary Evaluation and Scoring}
\label{sec:beval}

Given an evaluator LLM $E$, an input $x$ such as a source document, a transcript, or an instruction, an output $y$ such as a generated summary, a dialogue response, or a completion, and a binary question $q_i$, we define the \emph{binary evaluation function}
\[
 f_E(x, y, q_i) \in \{0, 1\},
\]
where $f_E(x, y, q_i) = 1$ if the evaluator answers ``yes'' and $0$ otherwise. Alongside each binary verdict, the evaluator produces a natural-language explanation $e_i$, enabling interpretability.

The per-dimension score for dimension $d$ is
\[
S_d(x, y) = \frac{1}{|\mathcal{Q}_d|} \sum_{q_i \in \mathcal{Q}_d} f_E(x, y, q_i).
\]
The overall score across all $N$ questions is
\[
S(x, y) = \frac{1}{N} \sum_{i=1}^{N} f_E(x, y, q_i).
\]
Both scores lie in $[0,1]$, where 1 indicates all criteria are satisfied.
To enable comparison with existing evaluation frameworks that use different scales, the scores can be mapped from $[0, 1]$ to any target interval $[a, b]$ via affine scaling:
\[
S'(x, y) = S(x, y) \cdot (b - a) + a.
\]

\subsection{Cross-Model Prompt Update}
\label{sec:evalopt}

\bindeval's binary question framework enables cross-model prompt update between evaluators. The key insight is that disagreements between a source evaluator and a target evaluator on specific binary questions provide a fine-grained signal for improvement: unlike holistic score differences, binary question disagreements identify exactly which criteria are being judged inconsistently across models. This makes it possible to use a stronger source model as a reference and iteratively update the prompt of a different, typically weaker, target model until its evaluator behavior matches the source more closely. Moreover, it is useful for updating prompts to maintain similar performance when migrating a model to a different family of models.

Let $E_{\text{src}}$ denote a source evaluator, treated as the reference model, and let $E_{\text{tgt}}$ denote a target evaluator whose prompt $P_E$ we wish to improve. Let $P_E^{(t)}$ denote the target evaluator's prompt at iteration $t$.

At each iteration $t$, the optimization proceeds in five steps:
\begin{enumerate}
    \item \textbf{Evaluate.} For each test case $(x_j, y_j)$, obtain binary evaluations from both models:
    \[
    \begin{aligned}
    A_j^{\text{src}} &= \{f_{E_{\text{src}}}(x_j, y_j, q_i)\}_{i=1}^{N}, \\
    A_j^{\text{tgt}} &= \{f_{E_{\text{tgt}}}(x_j, y_j, q_i; P_E^{(t-1)})\}_{i=1}^{N}.
    \end{aligned}
    \]
    \item \textbf{Identify disagreements.} Compute the set of questions on which the evaluators disagree:
    \[
    \Delta_j = \{q_i \in \mathcal{Q} : A_j^{\text{src}}(q_i) \neq A_j^{\text{tgt}}(q_i)\}.
    \]
    \item \textbf{Extract lessons.} A note-taker LLM $L_{\text{note}}$ analyzes each disagreement in context, extracting generalized lessons:
    \[
    \mathcal{L}_j = L_{\text{note}}(x_j, y_j, A_j^{\text{src}}, A_j^{\text{tgt}}, \Delta_j).
    \]
    \[
    \text{Dedup}(\ell_{\text{new}}, \mathcal{M}) =
\begin{cases}
\text{merge}(\ell_{\text{new}}, \ell_k), & \text{if } \ell_{\text{new}} \sim \ell_k \\ 
\text{add}(\ell_{\text{new}}), & \text{otherwise.}
\end{cases}
    \]
    The final set of unique lessons is $\mathcal{L}_{\text{unique}} = \text{Dedup}(\bigcup_j \mathcal{L}_j)$.
    \item \textbf{Update prompt.} For each unique lesson $\ell_k \in \mathcal{L}_{\text{unique}}$, an updater LLM identifies the relevant substring $s_k$ in the current prompt and produces a revised substring $s'_k$ that incorporates the lesson:
    \[
    P_E^{(t)} \leftarrow P_E^{(t)}.\text{replace}(s_k, s'_k).
    \]
\end{enumerate}

The loop terminates when the target evaluator's scores match the source evaluator's scores within a tolerance $\epsilon$ across all dimensions:
\[
|S_d^{\text{tgt},(t)} - S_d^{\text{src}}| < \epsilon \qquad \forall d \in \mathcal{D},
\]
or equivalently, when the target evaluator meets or exceeds the source evaluator on all dimensions.
The full algorithm is shown in Appendix~\ref{fig:algorithm}.

\subsection{Self Prompt Update}
\label{sec:genopt}

The same binary question framework can also be used for self prompt update in generation. Instead of aligning one evaluator to another model, this procedure iteratively improves a generator by using evaluator-identified failures as feedback on its own outputs. Given a generation LLM $L_{G}$ with prompt $P_G^{(t)}$ at iteration $t$:
\begin{enumerate}
    \item \textbf{Generate.} Produce outputs using the current prompt: $y_j^{(t)} = L_{G}(x_j; P_G^{(t)})$.
    \item \textbf{Evaluate.} Score each output using the potentially already-improved evaluator and collect failing questions:
    \[
    \mathcal{E}_j = \{(q_i, e_i) : f_E(x_j, y_j^{(t)}, q_i) = 0\},
    \]
    where $e_i$ is the evaluator's explanation for the failure.
    \item \textbf{Extract lessons.} A note-taker LLM analyzes the evaluation errors in context: $\mathcal{L}_j = L_{\text{note}}(x_j, y_j^{(t)}, \mathcal{E}_j)$.
    \item \textbf{Deduplicate and update.} Apply the same semantic deduplication and prompt rewriting procedure used for evaluator optimization, but now to $P_G$.
\end{enumerate}

The generation loop terminates when no evaluation errors remain or when the maximum number of iterations is reached.



\section{Experimental Setup}

We design two complementary sets of experiments. Part~I evaluates \bindeval's performance on established benchmarks with human annotations. Part~II demonstrates the iterative prompt-updating mechanism on both an unverifiable task and a verifiable task. Across these experiments, we use gpt-oss-120b and Claude Sonnet 4. To reduce randomness on LLM responses, we set the temperature to $0$ in all experiments and report the average over two runs.

\subsection{Metrics}

For evaluation quality, we report Spearman's rank correlation ($\rho$), Kendall's rank correlation ($\tau$), and Pearson correlation ($r$) between method scores and human judgments at the summary level. 

\subsection{Part I: Evaluation Quality Validation}

We follow the evaluation protocol of UniEval~\cite{zhong22} and evaluate on three established benchmarks.

\rwparagraph{SummEval}\cite{fabbri21}
A benchmark of 100 CNN/DM~\cite{see2017pointsummarizationpointergeneratornetworks} source articles, each summarized by 16 different summarization models, yielding 1,600 summary-level annotations. Human evaluators rated each summary on four dimensions: \emph{fluency}, \emph{coherence}, \emph{consistency}, and \emph{relevance}. Ratings are on a 1--5 Likert scale.

\rwparagraph{Topical-Chat}\cite{mehri20}
A benchmark of 60 dialogue responses generated by 6 dialogue models, annotated on six dimensions: \emph{naturalness}, \emph{coherence}, \emph{engagingness}, \emph{groundedness}, \emph{understandability}, and an \emph{overall} quality rating. Following Zhong et al.~\cite{zhong22}, we use four of these aspects.

\rwparagraph{QAGS}\cite{wang20qags}
A benchmark specifically targeting hallucination evaluation in summarization, comprising 235 samples from CNN/DM and 239 from XSum~\cite{narayan-etal-2018-dont}. Annotators rated the \emph{consistency} of each summary with respect to its source document.


\subsection{Part II: Iterative Prompt Updating}

We evaluate \bindeval's iterative prompt update mechanism (Algorithm~\ref{fig:algorithm}) on two tasks: \textbf{evaluator prompt optimization} on SummEval, which is unverifiable in the sense that there is no programmatic gold checker, and \textbf{generation prompt optimization} on IFBench~\cite{pyatkin2025generalizingverifiableinstructionfollowing}, which is verifiable via executable constraint checkers. For SummEval, we test two update modes: \emph{self-update}, where a single model (gpt-oss-120b) improves its own evaluator prompt using failures against human judgments, and \emph{cross-model update}, where a stronger model (Claude Sonnet 4) serves as the reference evaluator and lessons from disagreements are used to update the target model's prompt. See Appendix~\ref{app:exp_setup} for detailed experimental setups.



\section{Results}

\subsection{Evaluation Quality: SummEval}

\begin{table*}[!t]
\centering
\small
\caption{Summary-level Spearman $\rho$ / Kendall $\tau$ correlations on SummEval.}
\label{tab:summeval}
\begin{tabular}{lccccc}
\toprule
\textbf{Method} & \textbf{Coherence} & \textbf{Consistency} & \textbf{Fluency} & \textbf{Relevance} & \textbf{Average} \\
\midrule
ROUGE-1 & 0.167 / 0.126 & 0.160 / 0.130 & 0.115 / 0.094 & 0.326 / 0.252 & 0.192 / 0.150 \\
BERTScore & 0.284 / 0.211 & 0.110 / 0.090 & 0.193 / 0.158 & 0.312 / 0.243 & 0.225 / 0.175 \\
MoverScore & 0.159 / 0.118 & 0.157 / 0.127 & 0.129 / 0.105 & 0.318 / 0.244 & 0.191 / 0.148 \\
BARTScore & 0.448 / 0.342 & 0.382 / 0.315 & 0.356 / 0.292 & 0.356 / 0.273 & 0.385 / 0.305 \\
UniEval (T5) & 0.575 / 0.442 & 0.446 / 0.371 & 0.449 / 0.371 & 0.426 / 0.325 & 0.474 / 0.377 \\
G-Eval (GPT-4) & 0.582 / 0.457 & 0.507 / 0.425 & 0.506 / 0.455 & \textbf{0.547} /  \textbf{0.433} & 0.514 / 0.418 \\
{G-Eval (gpt-oss)} & 0.451 / 0.392 & 0.559 / 0.527 & 0.217 / 0.203 & 0.515 / 0.446 & 0.436 / 0.392 \\
UniEval (gpt-oss) & 0.237 / 0.208 & 0.489 / 0.476 & 0.000 / 0.000 & 0.288 / 0.256 & 0.254 / 0.235 \\
\textbf{\bindeval (gpt-oss)} & 0.523 / 0.448 & 0.585 / 0.548 & 0.252 / 0.235 & 0.428 / 0.366 & 0.447 / 0.399 \\
\textbf{\bindeval (Claude)} & \textbf{0.652 / 0.541} & \textbf{0.655 / 0.615} & \textbf{0.540 / 0.470} & 0.404 / 0.339 & \textbf{0.563 / 0.491}  \\
\bottomrule
\end{tabular}
\vspace{10pt}
\end{table*}

Table~\ref{tab:summeval} shows a clear ranking across evaluation paradigms. \bindeval (Claude) is the strongest method overall, achieving the best average Spearman and Kendall correlations and leading on coherence, consistency, and fluency. The largest gain is on consistency, where \bindeval reaches 0.655 / 0.615, suggesting that decomposing factual quality into multiple targeted checks is especially effective for summary evaluation. Relevance remains the main exception: G-Eval (GPT-4) is best on that dimension, indicating that some broader semantic judgments are still harder to capture with binary decomposition.

The additional gpt-oss runs clarify why decomposition matters. Under the same backbone, \bindeval (gpt-oss) outperforms both G-Eval (gpt-oss) and UniEval (gpt-oss) on average, driven by large gains on coherence and consistency. G-Eval with gpt-oss remains viable on numeric-scale dimensions such as consistency and relevance, but its fluency performance collapses. UniEval with gpt-oss is weaker still, with near-zero fluency correlation, showing that a single yes/no question is often too coarse for a general-purpose model. Overall, SummEval supports the core claim of the paper: multiple binary questions provide a more robust and transferable evaluation signal than either a single holistic score or a single Boolean judgment.

\begin{figure*}[!t]
\centering
\includegraphics[width=0.94\textwidth]{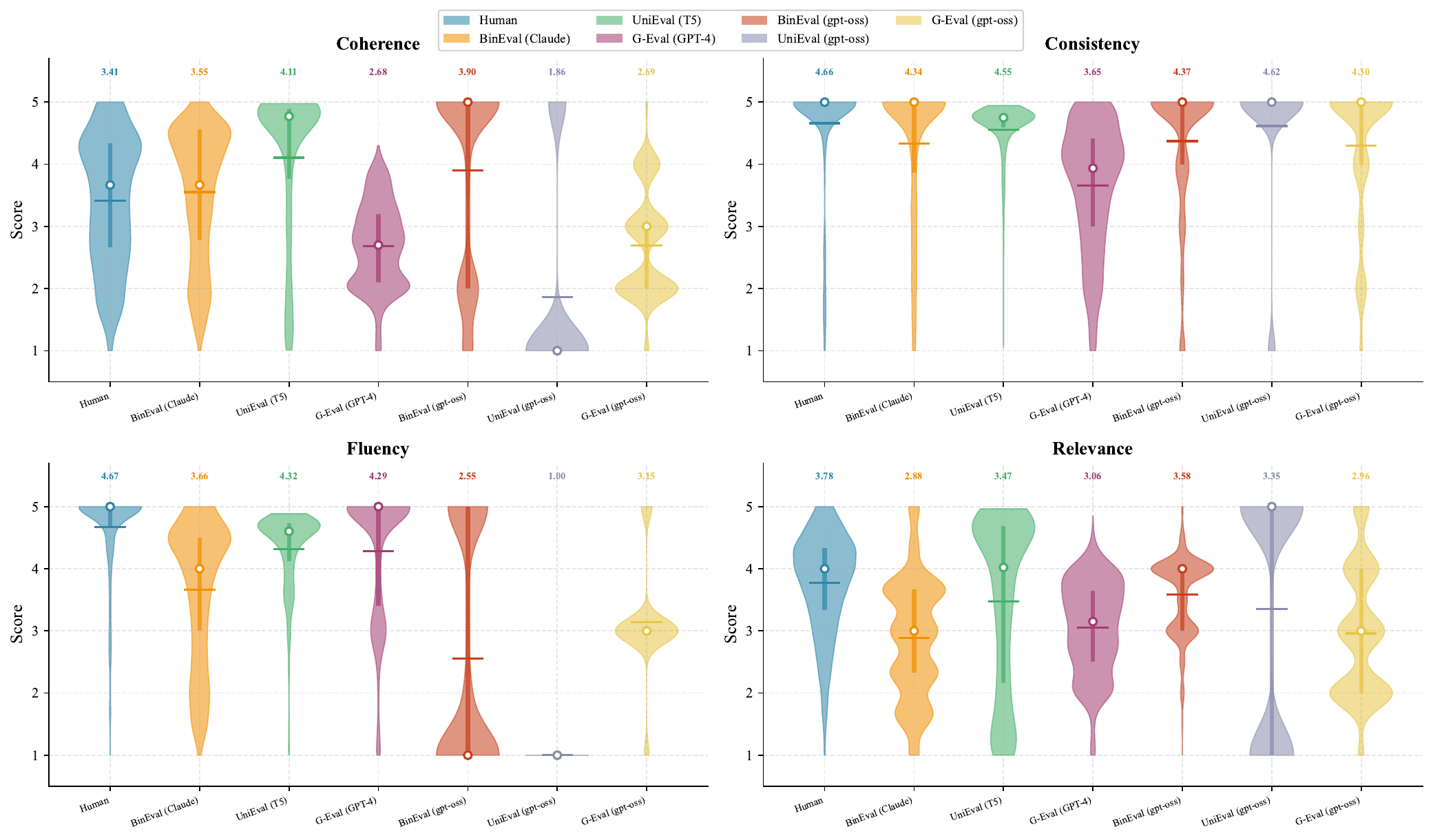}
\caption{Per-dimension score distributions on SummEval. \bindeval shows its strongest correlation on consistency. Its distribution is closest to the human shape while still preserving useful spread; it also remains competitive on coherence and fluency, even when its calibration is slightly more conservative than human ratings.}
\label{fig:violin-dim}
\end{figure*}

\begin{figure*}[!t]
\centering
\includegraphics[width=0.95\textwidth]{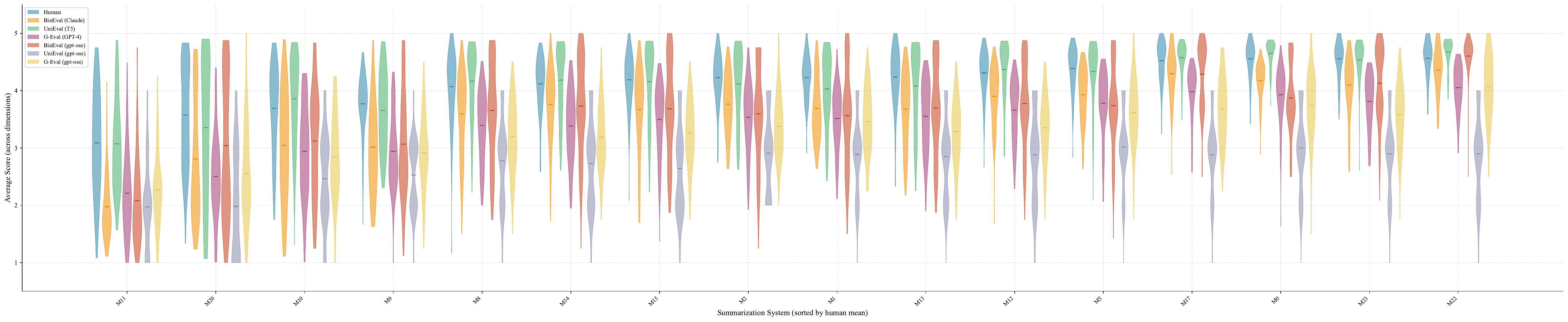}
\caption{Per-system average-score distributions on SummEval. Across the 16 summarization systems, \bindeval (Claude) best tracks the relative ordering of systems, while the weaker baselines produce flatter and less discriminative score patterns.}
\label{fig:violin-system}
\vspace{10pt}
\end{figure*}

\Cref{fig:violin-dim} gives a more nuanced view of these gains. The figure presents violin plots of score distributions on SummEval across four evaluation dimensions comparing human annotations with different methods. \bindeval is visually closest to the human distributions on consistency, where it largely matches the human concentration near the upper end while still retaining some low-scoring mass; this mirrors its largest correlation advantage in Table~\ref{tab:summeval}. Across dimensions, \bindeval (Claude) is generally among the methods most closely aligned with human judgments in central tendency and spread, with its strongest match on consistency. UniEval and G-Eval exhibit narrower, more concentrated distributions, suggesting weaker discrimination across systems. The gpt-oss-based variants consistently underestimate scores relative to humans, especially on coherence and relevance, where \bindeval (gpt-oss) and G-Eval (gpt-oss) show visibly lower means. Fluency is tightly clustered near the ceiling for all methods, reflecting the generally high fluency of modern summarization systems and the limited variance of this dimension. Notably, UniEval (gpt-oss) yields a near-degenerate fluency distribution, indicating its inability to differentiate quality along this axis. Overall, \bindeval's main strength is not perfect calibration on every dimension, but its ability to preserve meaningful relative variation, especially for factual consistency.

\Cref{fig:violin-system} provides the same comparison at the system level, where each score is averaged across the four SummEval dimensions and the 16 systems are ordered by ascending human mean. \bindeval (Claude) tracks the human ranking most faithfully, preserving the monotonic trend from weaker to stronger systems while maintaining visible separation among mid- and low-performing models. By contrast, UniEval and G-Eval exhibit more compressed score ranges that attenuate differences between systems, especially in the middle of the ranking. The gpt-oss-based methods are generally more conservative in absolute score level, but they still recover much of the broad system ordering. Another clear pattern is distributional width: \bindeval variants tend to show wider, more human-like within-system variance, whereas UniEval and G-Eval produce tighter violins that may understate genuine score variability. Agreement across methods is strongest for the highest human quality systems (rightmost), while the lower-quality systems show larger divergence, suggesting that distinguishing poor from  mediocre summaries remains a challenge for automated evaluation methods.

\subsection{Evaluation Quality: Topical-Chat}

The dialogue results show that \bindeval transfers effectively beyond summarization. \bindeval (Claude) achieves the best average Spearman correlation on Topical-Chat (0.632), with especially strong gains on naturalness and engagingness, while \bindeval (gpt-oss) remains competitive with G-Eval (gpt-oss) and substantially stronger than UniEval (gpt-oss). These results suggest that decomposing dialogue quality into multiple concrete questions is particularly helpful for subjective conversational criteria. Detailed results are provided in Appendix~\ref{app:topical_chat}.

\subsection{Evaluation Quality: QAGS}

QAGS highlights the advantage of decomposition most clearly. \bindeval (Claude) achieves the best average Spearman correlation (0.620), and even \bindeval (gpt-oss) substantially outperforms G-Eval (gpt-oss), whose binary prompt produces too little score granularity for reliable ranking. This suggests that decomposing factual consistency into several targeted questions is much more robust than relying on a single holistic or yes/no judgment, especially on hallucination-prone data such as XSum. Detailed results and discussion are provided in Appendix~\ref{app:qags}.

\subsection{Iterative Prompt Update}
\label{sec:prompt-update}

\subsubsection{SummEval: Evaluator Prompt Update}

Table~\ref{tab:prompt-update-summeval} reports test-set Spearman $\rho$ under iterative prompt update on the four SummEval dimensions. Both update modes improve three of the four dimensions. Self-update yields the largest single-dimension gain on fluency (+0.119), where the baseline prompt is especially weak and iterative refinement of both the evaluator rubric and the generated binary questions substantially improves alignment with human judgments. Cross-model update is strongest on consistency (+0.136), which is consistent with the idea that a stronger reference evaluator provides especially useful guidance for factual verification. Averaged across dimensions, self-update improves by +0.075, while cross-model update improves by +0.070.

Relevance resists improvement under both update modes. Inspecting the updated prompts suggests that lesson-driven refinements tend to over-decompose relevance into overly granular requirements, such as separate checks for every actor, motivation, and background event. These refinements make the evaluator more severe than human annotators rather than better aligned with them, which suggests that relevance remains a comparatively holistic judgment and is less amenable to fine-grained binary decomposition than dimensions with more concrete failure modes.

Three observations stand out. First, the two update modes are complementary: self-update helps most on coherence and fluency, while cross-model update helps most on consistency, indicating that human-score divergence and inter-model disagreement surface different classes of evaluator error. Second, most gains appear within the first one or two iterations; later iterations are more likely to degrade the prompt as lessons accumulate into competing instructions. Third, binary question regeneration is critical: the largest gains occur in iterations that alter not only the evaluator prompt but also the induced question decomposition, reinforcing that question design is itself a key lever for evaluation quality.

\begin{table}[!t]
\centering
\small
\caption{Evaluator prompt update on SummEval. Test-set Spearman $\rho$ with human judgments; $\Delta$ is the absolute improvement over the baseline. The best iteration is selected by early stopping on test performance.}
\label{tab:prompt-update-summeval}
\begin{adjustbox}{width=\columnwidth,center}
\begin{tabular}{lcccccc}
\toprule
& \multicolumn{3}{c}{\textbf{Self-Update}} & \multicolumn{3}{c}{\textbf{Cross-Model}} \\
\cmidrule(lr){2-4} \cmidrule(lr){5-7}
\textbf{Dimension} & Base & Best & $\Delta$ & Base & Best & $\Delta$ \\
\midrule
Coherence   & .521 & \textbf{.610} & +.089 & .524 & \textbf{.594} & +.070 \\
Consistency & .477 & \textbf{.568} & +.091 & .501 & \textbf{.637} & +.136 \\
Fluency     & .255 & \textbf{.375} & +.119 & .246 & \textbf{.318} & +.072 \\
Relevance   & .505 & .505          & .000  & .532 & .532          & .000  \\
\midrule
\textbf{Average} & .440 & \textbf{.515} & +.075 & .451 & \textbf{.520} & +.070 \\
\bottomrule
\end{tabular}
\end{adjustbox}
\vspace{10pt}
\end{table}

\subsubsection{IFBench: Generation Prompt Update}

Table~\ref{tab:prompt-update-ifbench} presents strict test-set accuracy on IFBench across prompt-update iterations. Self-update achieves a modest improvement, peaking at $38.0\%$ at iteration 3, which is a gain of $+3.4$ percentage points over its own iteration-0 baseline. However, the same run collapses by iteration 4, illustrating the fragility of repeated prompt rewriting. Cross-model update shows no improvement and in fact declines after the first update step, suggesting that the stronger judge's stricter standard can overcorrect the prompt rather than refine it.

The per-category breakdown in Table~\ref{tab:ifbench-category} reveals a sharp divide between \emph{promptable} and \emph{computational} constraints. Format and sentence constraints improve substantially, each by 17 percentage points, indicating that these tasks are often solved once the model is given clearer structural guidance. By contrast, count, ratio, words, and repeat constraints show little or no improvement. These constraints require precise computation during generation, such as maintaining counts, enforcing ratios, or filtering words by syllabic or lexical criteria. The extracted lessons often diagnose these failures correctly, but instructions such as ``maintain an internal counter'' do not endow the model with new computational ability. Instead, they accumulate into prompt bloat, which eventually harms even categories that were previously working well.

The main takeaway is that iterative prompt update is effective when the model already has the relevant capability but needs better guidance to express it. It is much less effective when failures reflect an underlying capability limitation rather than a prompting problem. In these cases, \bindeval still provides accurate diagnoses, but the resulting fixes are largely unactionable and can degrade performance through instruction overload.

\subsection{Case Study}
Appendix~\ref{app:case_study} presents both evaluation and prompt-update examples. It includes four SummEval case studies, one per dimension, showing that \bindeval can recognize coherence in a one-sentence summary, identify subtle factual errors, assign partial credit to garbled text, and separate incompleteness from irrelevance. The appendix also includes SummEval prompt-update examples for self-update and cross-model update, a relevance failure case where over-decomposition hurts alignment with human judgments, and an IFBench example highlighting the boundary between promptable failures and underlying computational limits. Together, these examples show that decomposition yields more justifiable scores and helps diagnose when prompt refinement succeeds or fails.

\begin{table}[!t]
\centering
\small
\caption{Generation prompt update on IFBench (test-set strict accuracy, \%).}
\label{tab:prompt-update-ifbench}
\begin{adjustbox}{width=\columnwidth,center}
\begin{tabular}{lccccc c}
\toprule
& \multicolumn{5}{c}{\textbf{Iteration}} & \\
\cmidrule(lr){2-6}
\textbf{Method} & 0 & 1 & 2 & 3 & 4 & \textbf{Peak} \\
\midrule
Self-update & 34.6 & 36.8 & 34.6 & \textbf{38.0} & 26.1 & \textbf{38.0} (+3.4) \\
Cross-model & \textbf{35.9} & 33.8 & --- & --- & --- & 35.9 (+0.0) \\
\midrule
No optimization (baseline) & \multicolumn{5}{c}{35.5} & 35.5 \\
\bottomrule
\end{tabular}
\end{adjustbox}
\end{table}

\begin{table}[!t]
\centering
\small
\caption{IFBench per-category accuracy (\%) under self-update.}
\label{tab:ifbench-category}
\begin{adjustbox}{width=\columnwidth,center}
\begin{tabular}{lcc l}
\toprule
\textbf{Category} & \textbf{Baseline} & \textbf{Peak} & \textbf{Trend} \\
\midrule
Format   & 52 & \textbf{69} & +17pp, responds to guidance \\
Sentence & 25 & \textbf{42} & +17pp, keyword and structure sensitive \\
Count    & 63 & 63          & degrades with more instructions \\
Ratio    & 22 & 22          & no change \\
Words    & 16 & 20          & marginal improvement \\
Repeat   & 17 & 17          & no change \\
\bottomrule
\end{tabular}
\end{adjustbox}
\vspace{10pt}
\end{table}

\subsection{Why Does Decomposition Work?}
\label{sec:analysis}

\emph{Why} does evaluating through multiple atomic binary questions outperform a single holistic judgment? We identify three contributing mechanisms and examine the evidence for each on SummEval (see Appendix~\ref{app:binary_questions} for the full question sets).

\rwparagraph{Complexity Reduction}
Each binary question isolates a single verifiable property, replacing one multi-faceted judgment with many simpler ones---mirroring the benefits of task decomposition in prompting~\cite{zhou22leasttomost,khot22decomposed}. A question like ``\emph{Are all named entities accurately represented?}'' is easier to answer reliably than ``\emph{Rate factual consistency from 1--5}.'' On consistency, the seven targeted questions yield yes-rates spread between 0.75 and 0.95 (\cref{tab:questions-consistency}), indicating each captures a distinct difficulty level. This pattern holds across dimensions: fluency, relevance, and coherence show yes-rate spreads of 0.48, 0.46, and 0.86 respectively.

\begin{figure}[!t]
\centering
\includegraphics[width=0.98\columnwidth]{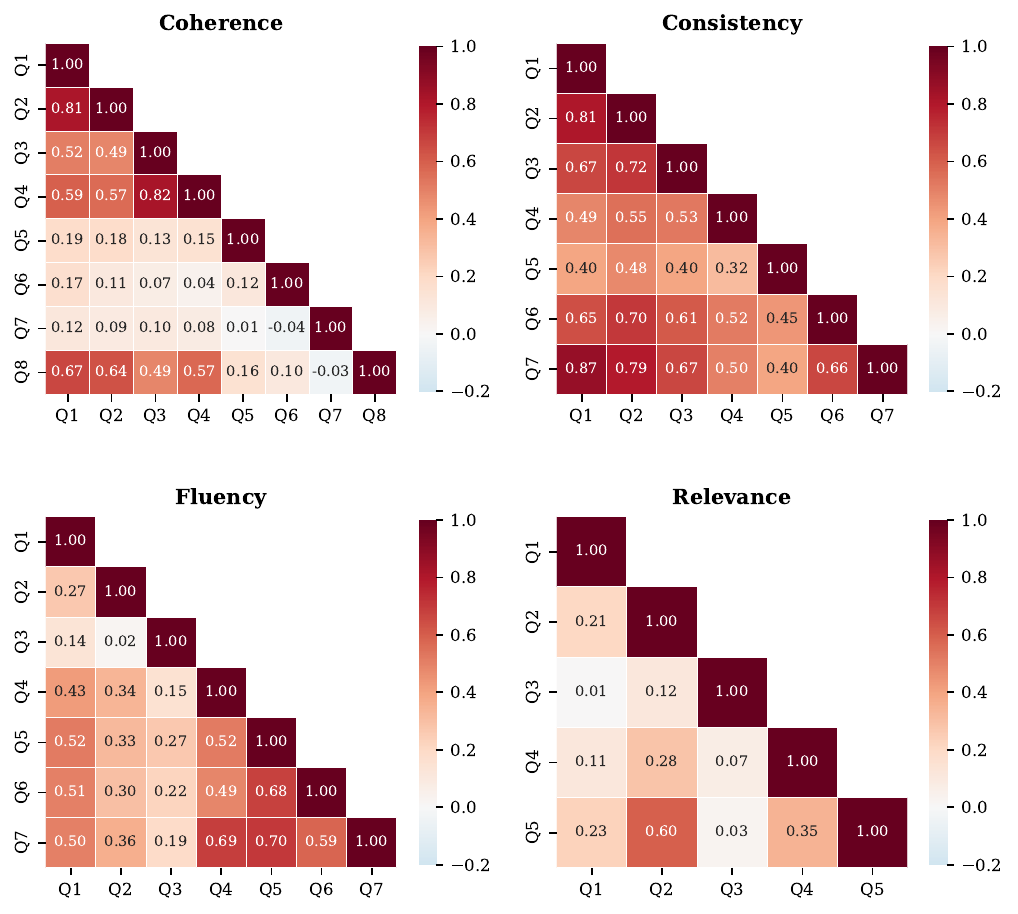}
\caption{Pairwise phi-coefficient correlation matrices within each SummEval dimension. Low off-diagonal values indicate questions capture distinct aspects of the dimension. Mean off-diagonal $\phi$ across all dimensions is $0.38$. See Appendix~\ref{app:binary_questions} for question definitions.}
\label{fig:correlation}
\vspace{10pt}
\end{figure}

\rwparagraph{Variance Reduction via Aggregation}
Aggregating $N$ weakly correlated binary classifiers reduces variance proportionally to $1/N$. \Cref{fig:correlation} shows this mechanism varies by dimension: relevance and coherence have the lowest mean inter-question correlations ($\phi = 0.20$ and $0.28$; 80\% and 64\% of pairs with $|\phi| < 0.3$), while fluency is moderate ($\phi = 0.39$; e.g., spelling Q2 vs.\ punctuation Q3 at $\phi = 0.02$). Consistency is the exception ($\phi = 0.58$, zero weak pairs), where questions like ``free of factual errors'' and ``no misrepresentation'' are inherently related ($\phi = 0.79$).

\rwparagraph{Coverage of Failure Modes}
Decomposition forces explicit enumeration of criteria, improving recall over holistic judgments. In fluency, spelling (Q2) and punctuation (Q3) are nearly uncorrelated ($\phi = 0.02$) with different yes-rates (0.71 vs.\ 0.33), catching disjoint failures. Relevance Q1 (main topic, 0.95) and Q3 (redundancy, 0.64) show $\phi = 0.01$. Consistency again is weakest: its least correlated pair has $\phi = 0.32$.

\rwparagraph{Dimension-Level Summary}
The three mechanisms contribute unequally. Relevance and coherence exhibit strong variance reduction and coverage. Fluency benefits from all three. Consistency is the most instructive: weakest variance reduction and coverage, yet the largest gain over UniEval (+0.195 Spearman $\rho$), suggesting that complexity reduction alone---decomposing factual verification into targeted sub-checks---can be the dominant driver. From a practical standpoint, practitioners can inspect generated questions for these properties (yes-rate spread, inter-question correlation, pairwise coverage) to anticipate where decomposition will help most and where refinement is needed.

\section{Discussion}

\rwparagraph{Failure Modes}
Decomposition works best for concrete criteria such as factual consistency, where errors can be tied to specific claims or entities and can therefore be checked with relatively clear yes/no decisions. It is less reliable for subjective qualities, where human judgments are more holistic and less reducible to a set of binary checks. In such cases, the quality of the evaluation depends heavily on whether the generated questions capture the aspects that humans actually weigh when forming an overall judgment. The appendix shows both patterns: relevance can degrade when decomposition becomes too strict, and prompt update helps less when failures reflect the model's base capability rather than its instructions. On IFBench, clearer prompts help with format and sentence-level constraints but not with counting or ratio tracking, suggesting that some errors stem from execution limits rather than task specification alone.

\rwparagraph{Computational Cost}
\bindeval trades efficiency for diagnostic value. Compared with a single holistic judgment, it must generate binary questions and answer each of them. This increases both the number of model calls and the total amount of text processed during evaluation. Prompt updating adds note-taking, lesson deduplication, and meta-prompt rewriting, though batching keeps the first two modest and prompt rewriting is shared by most update methods. The main recurring cost is question-level evaluation.

\rwparagraph{Limitations}
The method still depends on question quality: if important criteria are missing, the final score will miss them. It also assumes that the fraction of satisfied questions maps approximately linearly to overall quality, which need not always hold.

\subsection{Decomposed Evaluation vs.\ Holistic Scoring}
\label{sec:decomposed-vs-holistic}

\begin{figure*}[t]
\begin{tcolorbox}[casebox,
  title={Consistency Example: Why question decomposition outperforms holistic scoring}]

\small

\textbf{Task:} Evaluate the \emph{factual consistency} of a summary against its source article.

\medskip

\textbf{Source article} (excerpt): \textit{%
``The U.S.\ military said a Russian SU-27 Flanker conducted a barrel roll over
a U.S.\ RC-135U \ldots\ The Pentagon called the intercept \emph{unsafe and
unprofessional} \ldots\ The Russian Defense Ministry said the jet was
scrambled to identify the aircraft \ldots''}

\medskip

\textbf{System summary:} \textit{%
``The U.S.\ RC-135U was flying over the Baltic Sea when it was intercepted by
a Russian SU-27 Flanker. \underline{The Pentagon said} the Russian jet flew
around the U.S.\ plane several times \underline{to identify it and get its
tail number}. Read more:
\underline{http://dailycaller.com/2017/10/29/\ldots}''}

\medskip

\begin{adjustbox}{max width=\linewidth}
\begin{tabular}{@{}l c c p{7.8cm}@{}}
\toprule
\textbf{Method} & \textbf{Score} & \textbf{$|\Delta|$} & \textbf{Why} \\
\midrule
Human & \textbf{2.0} & --- & Contains multiple factual errors \\[2pt]
\rowcolor{clBinClaude!8}
\textbf{BinEval (Claude)} & \textbf{1.57} & \textcolor{clGreen}{\textbf{0.43}}
  & Catches 4 of 7 errors via decomposed questions \\[2pt]
BinEval (gpt-oss) & 3.0 & 1.0
  & Only catches 2 of 7 errors; misses misattribution \\[2pt]
G-Eval (gpt-oss & 5.0 & \textcolor{clRed}{\textbf{3.0}}
  & Single holistic score: ``looks consistent'' \\[2pt]
UniEval (gpt-oss) & 5.0 & \textcolor{clRed}{\textbf{3.0}}
  & Single Q ``Is this consistent?''\ $\to$ \yes \\
\bottomrule
\end{tabular}
\end{adjustbox}

\medskip

\textbf{BinEval (Claude) — decomposed evaluation:}

\vspace{2pt}
\begin{adjustbox}{max width=\linewidth}
\begin{tabular}{@{}c p{0.44\linewidth} c p{0.44\linewidth}@{}}
\textcolor{clRed}{\small$\boldsymbol\times$} &
  \textbf{Q1.}\ Claims supported? \no\ --- Pentagon's statement
  is misattributed (Russia said ``identify,'' not the Pentagon). &
\textcolor{clGreen}{\small$\boldsymbol\checkmark$} &
  \textbf{Q4.}\ Numbers correct? \yes\ --- Aircraft types
  (RC-135U, SU-27) match the source. \\[4pt]
\textcolor{clRed}{\small$\boldsymbol\times$} &
  \textbf{Q2.}\ No fabrication? \no\ --- URL
  \texttt{dailycaller.com/\ldots} not appear in the source. &
\textcolor{clRed}{\small$\boldsymbol\times$} &
  \textbf{Q5.}\ Causal relations preserved? \no\ --- Conflates Pentagon
  and Russian accounts of the intercept. \\[4pt]
\textcolor{clRed}{\small$\boldsymbol\times$} &
  \textbf{Q3.}\ Entities accurate? \no\ --- Misattributes
  Russia's stated purpose to the Pentagon. &
\textcolor{clGreen}{\small$\boldsymbol\checkmark$} &
  \textbf{Q6--7.}\ No hallucinations?\ No misrepresentation of scope?
  \yes\ --- Core event is described. \\
\end{tabular}
\end{adjustbox}

\vspace{4pt}
\textbf{Score:} $3/7$ questions answered \yes\ $\;\Rightarrow\; 3/7 \approx 0.43
\;\xrightarrow{\text{scale to 1--5}}\; \mathbf{1.57}$
\quad (human:\ \textbf{2.0},\ error:\ 0.43)

\medskip
\hrule
\vspace{4pt}

\textbf{Key insight.}\
G-Eval and UniEval assign perfect consistency (5.0) because the summary
\emph{looks plausible} at a surface level --- it mentions the correct aircraft
and event.
BinEval's decomposed questions probe each factual claim independently,
catching the misattribution (Q1, Q3), fabricated URL (Q2), and conflated
accounts (Q5).
The resulting $3/7$ score closely matches the human rating of~2.0,
while holistic methods miss every error.

\end{tcolorbox}
\caption{%
Illustrative SummEval consistency example.
The summary contains subtle factual errors (underlined) that holistic scoring
methods miss.
BinEval decomposes consistency into seven binary questions, each targeting
a specific error type, producing a score closely aligned with the human
judgment.
}
\label{fig:bineval-example-consistency}
\vspace{15pt}
\end{figure*}

Figure~\ref{fig:bineval-example-consistency} illustrates a representative failure mode of holistic evaluation methods. The summary under evaluation contains three distinct factual errors (underlined): a misattribution of Russia's stated purpose to the Pentagon, a fabricated external URL absent from the source, and a conflation of the two parties' accounts of the intercept. Despite these errors, both G-Eval and UniEval assign a perfect consistency score of 5.0, because the summary is \emph{surface-plausible}---it names the correct aircraft types and describes the general event accurately. Holistic scoring conflates local correctness with global consistency, rewarding fluent, topically coherent text even when specific claims are wrong.

BinEval avoids this by decomposing consistency into seven targeted binary questions, each probing a distinct claim type: factual support, fabrication, entity accuracy, numerical correctness, causal fidelity, hallucination, and scope representation. Questions Q1, Q3, and Q5 directly surface the misattribution and conflation; Q2 flags the fabricated URL. The resulting score of $3/7 \approx 1.57$ (scaled to 1--5) closely matches the human rating of 2.0 ($|\Delta| = 0.43$), whereas G-Eval and UniEval diverge by 3.0 points. 
This example motivates the core design principle of BinEval: fine-grained binary questions act as \emph{claim-level probes}, making errors visible that aggregate scoring systematically obscures.
Critically, this granularity also makes the feedback actionable, as each failed question directly identifies the error type, enabling targeted corrections to either the summarizer or the evaluator prompt.


\section{Conclusion}

We presented \textsc{BinEval}, a task-agnostic, training-free framework that evaluates LLM outputs by decomposing criteria into atomic binary questions. Across SummEval, Topical-Chat, and QAGS, it matches or outperforms strong evaluators while also supporting iterative prompt optimization on summarization and IFBench. Because each score is grounded in individual verdicts with explanations, \textsc{BinEval} offers interpretable feedback that helps practitioners diagnose and improve LLM systems, and suggests atomic binary decomposition as a promising direction for broader evaluation tasks. These results indicate that interpretability and strong evaluation performance need not come at the expense of scalability or flexibility. Looking ahead, we see natural extensions to agentic and multi-turn settings, where fine-grained, claim-level feedback is especially valuable for identifying where and why a system goes wrong.



\section*{Impact Statement}

This paper presents work whose goal is to advance the field of machine learning through more interpretable and scalable evaluation of language model outputs. There are many potential societal consequences of our work, including the possibility of improving the reliability of automated evaluation pipelines used in research and deployment. At the same time, evaluator models can inherit the biases and blind spots of the underlying language models used to instantiate them, so any deployment of \bindeval should be paired with human oversight in high-stakes settings.

\bibliography{example_paper}
\bibliographystyle{icml2026}

\appendix
\onecolumn

\section{Case Study}
\label{app:case_study}

\subsection{Effective Evaluation: Illustrative Examples}
\label{sec:evaluation-examples}

\newcommand{\scorechart}[9]{%
  \begin{tikzpicture}[
      y=0.38cm,
      every node/.style={font=\scriptsize},
    ]
    \pgfmathsetmacro{\unitw}{\linewidth / #3 * 0.50}
    \pgfmathsetmacro{\xmax}{#3}
    \pgfmathsetmacro{\barh}{0.30}
    \def\rowUEG{0} \def\rowUET{1} \def\rowGEG{2}
    \def\rowGE4{3} \def\rowBG{4}  \def\rowBC{5} \def\rowH{6}
    \fill[clUniEvalGPT!80] (0,\rowUEG) rectangle ({#9*\unitw pt},\rowUEG+\barh);
    \fill[clUniEvalT5!80]  (0,\rowUET) rectangle ({#8*\unitw pt},\rowUET+\barh);
    \fill[clGEvalGPT!80]   (0,\rowGEG) rectangle ({#7*\unitw pt},\rowGEG+\barh);
    \fill[clGEvalGPT4!80]  (0,\rowGE4) rectangle ({#6*\unitw pt},\rowGE4+\barh);
    \fill[clBinGPT!80]     (0,\rowBG)  rectangle ({#5*\unitw pt},\rowBG+\barh);
    \fill[clBinClaude!80]  (0,\rowBC)  rectangle ({#4*\unitw pt},\rowBC+\barh);
    \fill[clHuman!80]      (0,\rowH)   rectangle ({#2*\unitw pt},\rowH+\barh);
    \draw[clHuman, densely dashed, line width=0.7pt]
      ({#2*\unitw pt},-0.3) -- ({#2*\unitw pt},\rowH+0.5);
    \node[anchor=east] at (-0.06cm,\rowH+\barh/2)   {Human};
    \node[anchor=east] at (-0.06cm,\rowBC+\barh/2)  {\textbf{BinEval\,(Cl.)}};
    \node[anchor=east] at (-0.06cm,\rowBG+\barh/2)  {BinEval\,(gpt-oss)};
    \node[anchor=east] at (-0.06cm,\rowGE4+\barh/2) {G-Eval\,(GPT-4)};
    \node[anchor=east] at (-0.06cm,\rowGEG+\barh/2) {G-Eval\,(gpt-oss)};
    \node[anchor=east] at (-0.06cm,\rowUET+\barh/2) {UniEval\,(T5)};
    \node[anchor=east] at (-0.06cm,\rowUEG+\barh/2) {UniEval\,(gpt-oss)};
    \node[anchor=west,font=\scriptsize\bfseries] at ({#2*\unitw pt+0.06cm},\rowH+\barh/2)   {#2};
    \node[anchor=west] at ({#4*\unitw pt+0.06cm},\rowBC+\barh/2)  {#4};
    \node[anchor=west] at ({#5*\unitw pt+0.06cm},\rowBG+\barh/2)  {#5};
    \node[anchor=west] at ({#6*\unitw pt+0.06cm},\rowGE4+\barh/2) {#6};
    \node[anchor=west] at ({#7*\unitw pt+0.06cm},\rowGEG+\barh/2) {#7};
    \node[anchor=west] at ({#8*\unitw pt+0.06cm},\rowUET+\barh/2) {#8};
    \node[anchor=west] at ({#9*\unitw pt+0.06cm},\rowUEG+\barh/2) {#9};
    \draw[->] (0,-0.4) -- ({\xmax*\unitw pt+0.2cm},-0.4);
    \foreach \v in {0,1,...,\xmax}{
      \draw ({\v*\unitw pt},-0.4) -- ({\v*\unitw pt},-0.52) node[below]{\v};
    }
    \node[anchor=south, font=\small\bfseries] at ({\xmax/2*\unitw pt}, \rowH+0.65) {#1};
  \end{tikzpicture}%
}


\begin{figure*}[!h]
\begin{tcolorbox}[casebox, title={(a) Coherence --- Single-sentence summary}]
\small

\textbf{Summary:} \textit{``Speed camera has been turned round and is pointing at this house in Birmingham, West Midlands.''}

\smallskip
\begin{adjustbox}{max width=\linewidth}
\begin{tabular}{@{}l c c p{6.5cm}@{}}
\toprule
\textbf{Method} & \textbf{Score} & \textbf{$|\Delta|$} & \textbf{Explanation} \\
\midrule
Human & \textbf{4.67} & --- & Coherent: clear and on-topic \\
\rowcolor{clBinClaude!8}
\textbf{BinEval (Claude)} & \textbf{4.56} & \textcolor{clGreen}{\textbf{0.11}} & 7/8 Qs \yes{} --- trivially coherent as single sentence \\
G-Eval (gpt-oss) & 1.00 & \textcolor{clRed}{3.67} & Penalises brevity as ``incoherent'' \\
UniEval (gpt-oss) & 1.00 & \textcolor{clRed}{3.67} & ``Is this coherent?'' $\to$ \no{} \\
\bottomrule
\end{tabular}
\end{adjustbox}

\smallskip
\textbf{Decomposed reasoning:}\;
\yes\,Q1 (structured) \;
\yes\,Q2 (logical order) \;
\yes\,Q3 (transitions) \;
\yes\,Q4 (no repetition) \;
\yes\,Q5 (unified focus) \;
\yes\,Q6 (main topic) \;
\no\,Q7 (misses some details) \;
\yes\,Q8 (no contradictions)

\smallskip
\textbf{Insight:}
A single sentence \emph{trivially} satisfies ordering, non-contradiction, and focus criteria.
The one \no{} (incomplete coverage) yields a proportional penalty: $7/8 \to 4.56$, closely matching the human score.
Holistic methods conflate \emph{completeness} with \emph{coherence}, assigning the minimum score.
\end{tcolorbox}


\begin{tcolorbox}[casebox, title={(b) Consistency --- Subtle factual errors in a plausible summary}]
\small

\textbf{Source} (excerpt): \textit{``The Pentagon called the intercept \emph{unsafe and unprofessional}\ldots\ The Russian Defense Ministry said the jet was scrambled \emph{to identify} the aircraft\ldots''}

\smallskip
\textbf{Summary:} \textit{``The U.S.\ RC-135U was flying over the Baltic Sea when it was intercepted by a Russian SU-27 Flanker. \underline{The Pentagon said} the Russian jet flew around the plane \underline{to identify it}. Read more: \underline{http://dailycaller.com/\ldots}''}

\smallskip
\begin{adjustbox}{max width=\linewidth}
\begin{tabular}{@{}l c c p{6.5cm}@{}}
\toprule
\textbf{Method} & \textbf{Score} & \textbf{$|\Delta|$} & \textbf{Explanation} \\
\midrule
Human & \textbf{2.00} & --- & Multiple factual errors \\
\rowcolor{clBinClaude!8}
\textbf{BinEval (Claude)} & \textbf{1.57} & \textcolor{clGreen}{\textbf{0.43}} & 3/7 Qs \yes{} --- catches misattribution, URL, conflation \\
G-Eval (gpt-oss) & 5.00 & \textcolor{clRed}{3.00} & ``Looks consistent'' at surface level \\
UniEval (gpt-oss) & 5.00 & \textcolor{clRed}{3.00} & ``Is this consistent?'' $\to$ \yes{} \\
\bottomrule
\end{tabular}
\end{adjustbox}

\smallskip
\textbf{Decomposed reasoning:}\;
\no\,Q1 (misattributes Russia's purpose to Pentagon) \;
\no\,Q2 (fabricated URL) \;
\no\,Q3 (wrong entity role) \;
\yes\,Q4 (aircraft types correct) \;
\no\,Q5 (conflates Pentagon/Russian accounts) \;
\yes\,Q6--7 (core event described)

\smallskip
\textbf{Insight:}
The summary mentions correct entities (RC-135U, SU-27) and describes the real event, so holistic methods see it as consistent.
BinEval's decomposed questions probe \emph{each claim independently}, catching the misattribution (\no\,Q1, Q3), fabricated URL (\no\,Q2), and conflated accounts (\no\,Q5). Score: $3/7 \to 1.57$, close to human 2.0.
\end{tcolorbox}


\begin{tcolorbox}[casebox, title={(c) Fluency --- Garbled summary with partial readability}]
\small

\textbf{Summary:} \textit{``\,`Space invaders' was developed in japan back in 1970. Japanese can sleep soundly in their beds tonight as government's top military official. He also fought muhammad ali in 1976. Inoki has appeared in the u.s.-based wwe.''}

\smallskip
\begin{adjustbox}{max width=\linewidth}
\begin{tabular}{@{}l c c p{6.5cm}@{}}
\toprule
\textbf{Method} & \textbf{Score} & \textbf{$|\Delta|$} & \textbf{Explanation} \\
\midrule
Human & \textbf{2.00} & --- & Some errors but partially readable \\
\rowcolor{clBinClaude!8}
\textbf{BinEval (Claude)} & \textbf{1.50} & \textcolor{clGreen}{\textbf{0.50}} & 2/8 Qs \yes{} --- recognises partial readability \\
G-Eval (gpt-oss) & 1.00 & 1.00 & Minimum score: ``poor quality'' \\
UniEval (gpt-oss) & 1.00 & 1.00 & ``Is this fluent?'' $\to$ \no{} \\
\bottomrule
\end{tabular}
\end{adjustbox}

\smallskip
\textbf{Decomposed reasoning:}\;
\no\,Q1 (sentence fragment: ``as government's top military official'') \;
\yes\,Q2 (no spelling errors) \;
\no\,Q3 (punctuation: backtick quotes, missing caps) \;
\no\,Q4 (imprecise: ``1970'' vs ``late 1970s'') \;
\no\,Q5 (run-on fragment in sentence 2) \;
\no\,Q6 (unnatural jumps between topics) \;
\no\,Q7 (requires re-reading) \;
\yes\,Q8 (main points still comprehensible)

\smallskip
\textbf{Insight:}
Human annotators rate this 2/3 (not the worst) because the text is partially readable despite errors.
BinEval captures this nuance: Q2 (\yes, no spelling errors) and Q8 (\yes, comprehensible gist) prevent a floor score. Score: $2/8 \to 1.50$, between 1 and 2.
G-Eval and UniEval assign the minimum because any fluency issue triggers a blanket negative judgment.
\end{tcolorbox}

\caption{Four illustrative SummEval examples, one per evaluation dimension.
In each case, BinEval's question decomposition produces scores closely aligned
with human judgments by independently assessing multiple quality facets.
Holistic methods (G-Eval, UniEval with gpt-oss) collapse to extreme scores on
edge cases---short-but-correct summaries, partially readable text, or concise
one-liners---because a single judgment conflates orthogonal quality dimensions.}
\label{fig:bineval-four-cases}
\end{figure*}


\begin{figure}[thp]
\begin{tcolorbox}[casebox, title={(d) Relevance --- Concise but topically relevant one-liner}]
\small

\textbf{Source} (excerpt): \textit{``ISIS released more than 200 Yazidis\ldots\ mostly women, children, and elderly\ldots\ A senior Peshmerga commander said they were released in groups\ldots\ The freed captives appeared very tired\ldots''}

\smallskip
\textbf{Summary:} \textit{``ISIS released over 200 Yazidis on Wednesday.''}

\smallskip
\begin{adjustbox}{max width=\linewidth}
\begin{tabular}{@{}l c c p{6.5cm}@{}}
\toprule
\textbf{Method} & \textbf{Score} & \textbf{$|\Delta|$} & \textbf{Explanation} \\
\midrule
Human & \textbf{3.33} & --- & Relevant but incomplete \\
\rowcolor{clBinClaude!8}
\textbf{BinEval (Claude)} & \textbf{3.67} & \textcolor{clGreen}{\textbf{0.33}} & 4/6 Qs \yes{} --- on-topic, no padding, but sparse \\
G-Eval (gpt-oss) & 1.00 & \textcolor{clRed}{2.33} & Penalises brevity as ``irrelevant'' \\
UniEval (gpt-oss) & 1.00 & \textcolor{clRed}{2.33} & ``Is this relevant?'' $\to$ \no{} \\
\bottomrule
\end{tabular}
\end{adjustbox}

\smallskip
\textbf{Decomposed reasoning:}\;
\no\,Q1 (omits key details: demographics, conditions) \;
\yes\,Q2 (no fabricated content) \;
\yes\,Q3 (no redundancy) \;
\yes\,Q4 (no trivial padding) \;
\no\,Q5 (too sparse, misses important aspects) \;
\yes\,Q6 (included content is relevant)

\smallskip
\textbf{Insight:}
The summary captures the central event accurately but is too brief.
BinEval rewards what it \emph{does} right (on-topic, no fabrication, no padding) while penalising omissions (Q1, Q5). Score: $4/6 \to 3.67$, matching human 3.33.
G-Eval and UniEval again conflate \emph{incompleteness} with \emph{irrelevance}, assigning the minimum score despite the summary being factually on-topic.
\end{tcolorbox}
\end{figure}


\begin{figure}[h]
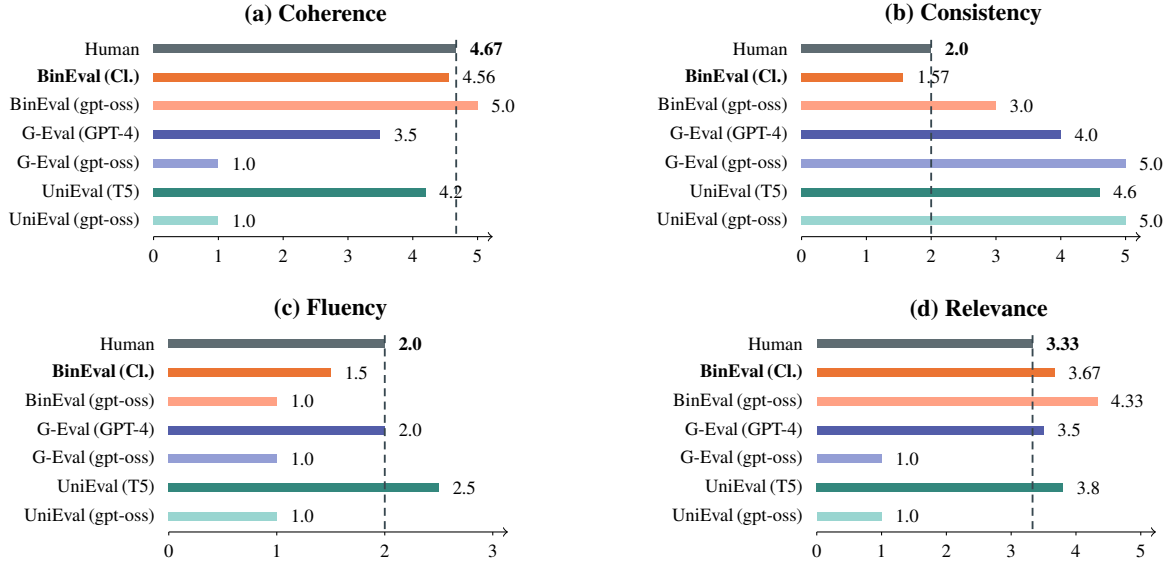

\centering

\begin{minipage}[t]{0.5\linewidth}\centering
\scorechart{(a) Coherence}{4.67}{5}{4.56}{5.0}{3.5}{1.0}{4.2}{1.0}
\end{minipage}\hfill
\begin{minipage}[t]{0.5\linewidth}\centering
\scorechart{(b) Consistency}{2.0}{5}{1.57}{3.0}{4.0}{5.0}{4.6}{5.0}
\end{minipage}

\vspace{6pt}

\begin{minipage}[t]{0.5\linewidth}\centering
\scorechart{(c) Fluency}{2.0}{3}{1.5}{1.0}{2.0}{1.0}{2.5}{1.0}
\end{minipage}\hfill
\begin{minipage}[t]{0.5\linewidth}\centering
\scorechart{(d) Relevance}{3.33}{5}{3.67}{4.33}{3.5}{1.0}{3.8}{1.0}
\end{minipage}

\caption{Score comparisons for four illustrative SummEval examples, one per dimension.
Dashed line marks the human reference.
BinEval (Claude) consistently tracks human scores across all dimensions.
G-Eval (GPT-4) and UniEval (T5) --- the published baselines --- perform reasonably,
but when their evaluation formats are applied to gpt-oss without Monte Carlo sampling
or fine-tuning, scores collapse on edge cases.}
\label{fig:bineval-four-examples-bars}
\end{figure}

\pagebreak
\subsection{Prompt Evolution: Illustrative Examples}
\label{sec:prompt-evolution-examples}

This section illustrates how BinEval's iterative prompt update modifies evaluation and generation prompts across iterations, with examples of both successful updates and failure modes.

\subsubsection{Example 1: Self-Update on Coherence (SummEval)}
\label{sec:example-self-coherence}

\textbf{Result:} Spearman $\rho$ improved from .521 (baseline) to .610 (iteration 1).

The self-update pipeline identified that the baseline coherence prompt was \emph{too strict} on single-sentence summaries and penalized omission of background details, while human annotators focused primarily on logical flow. Three representative lessons were extracted:

\begin{tcolorbox}[colback=gray!5, colframe=orange!80, title={Extracted Lessons (Self-Update, Coherence)}, fonttitle=\small\bfseries]
\small
\begin{enumerate}[nosep]
  \item \textbf{Implicit transitions are acceptable.} Require logical connections but do not demand explicit cue words (``because,'' ``therefore''). Implicit continuity suffices if the narrative flows.
  \item \textbf{Add a central-claim relevance criterion.} Each sentence should advance the article's main claim; sentences that do not contribute are non-contributory regardless of grammatical correctness.
  \item \textbf{Do not penalize omission of background details.} Missing context should not lower coherence as long as the core fact and conflict remain clear.
\end{enumerate}
\end{tcolorbox}

\noindent These lessons produced targeted edits to the evaluation rubric:

\begin{table}[h]
\centering
\small
\caption{Coherence prompt: key changes from iteration 0 to iteration 1.}
\label{tab:coh-diff}
\begin{tabular}{p{0.45\textwidth} p{0.45\textwidth}}
\toprule
\textbf{Iteration 0 (Baseline)} & \textbf{Iteration 1 (Updated)} \\
\midrule
\textit{``...logical connections between sentences (explicit cues like `because,' `therefore,' or implicit continuity)...''} & \textit{``...logical connections between sentences (\textbf{implicit connections are acceptable}; explicit markers like `after,' `because' are helpful but \textbf{not required})...''} \\[6pt]
\textit{``...global focus --- every sentence stays directly related to the main topic.''} & \textit{``...\textbf{relevance to the central claim}: identify the article's main claim or core fact and ensure every sentence \textbf{advances or supports that claim}. If a sentence does not advance the main argument, treat it as non-contributory.''} \\[6pt]
\textit{``Penalize only for poor logical flow, redundancy, misordering, or off-topic content, not for missing facts.''} & \textit{``\textbf{Do not penalize the summary for omitting background details} as long as the essential conflict or core fact remains clear.''} \\
\bottomrule
\end{tabular}
\end{table}

\noindent \textbf{Why it works:} The lessons correctly identified a systematic bias---the model over-penalized brevity---and the updated rubric explicitly instructs the evaluator to tolerate omissions while adding a concrete ``central claim'' criterion that better aligns with how human annotators judge coherence.

\subsubsection{Example 2: Cross-Model Update on Consistency (SummEval)}
\label{sec:example-cross-consistency}

\textbf{Result:} Spearman $\rho$ improved from .501 (baseline) to .637 (iteration 1).

Claude (source evaluator) correctly distinguished between \emph{omission} (not mentioning a source fact) and \emph{contradiction} (stating something unsupported). gpt-oss (target) conflated these, penalizing summaries that simply omitted details. Key disagreement-driven lessons:

\begin{tcolorbox}[colback=blue!3, colframe=blue!40, title={Extracted Lessons (Cross-Model, Consistency)}, fonttitle=\small\bfseries]
\small
\begin{enumerate}[nosep]
  \item \textbf{Omission $\neq$ inconsistency.} A summary that omits details from the source is not factually inconsistent; only statements \emph{present in the summary} that are unsupported should be penalized.
  \item \textbf{Semantic equivalence via arithmetic.} Converting ``83rd minute'' to ``seven minutes remaining'' (in a 90-minute match) is a valid transformation, not a hallucination.
  \item \textbf{Subject--role misattribution.} When summaries restructure clauses, verify that entities are attached to the correct verbs (e.g., ``X restarted his row with Z'' misattributes if the source says ``X had a row with Y and drew 0--0 with Z'').
\end{enumerate}
\end{tcolorbox}

\noindent The updated prompt grew substantially (from 4 evaluation steps to 6, with detailed guidance on literal interpretation, subject verification, and semantic equivalence). The critical structural addition:

\begin{tcolorbox}[colback=white, colframe=black!30, fontupper=\small\ttfamily]
\textbf{Added to Evaluation Steps (cross-model, iteration 1):}\\[3pt]
"For each statement in the summary, check whether it is supported by the article. \textit{The summary does not need to cover all details from the article. Omitting information is not a factual error.} Only flag statements that are present in the summary but are unsupported or contradicted."
\end{tcolorbox}

\noindent \textbf{Why it works:} The cross-model signal pinpointed a fundamental conceptual error (conflating omission with contradiction) that human-score divergence alone could not have surfaced so clearly. The +.136 improvement---the largest in our experiments---demonstrates that inter-model disagreement can identify systematic evaluation biases that self-reflection misses.

\subsubsection{Example 3: Failure Case --- Relevance (SummEval)}
\label{sec:example-failure-relevance}

\textbf{Result:} Spearman $\rho$ \emph{decreased} from .505 to .357 after applying lessons.

The self-update pipeline correctly diagnosed that the model was too lenient on relevance---giving perfect scores to summaries that captured the headline fact but omitted key actors and motivations. However, the fix made the prompt \emph{too strict}:

\begin{tcolorbox}[colback=red!3, colframe=red!40, title={Extracted Lessons (Self-Update, Relevance --- Led to Degradation)}, fonttitle=\small\bfseries]
\small
\begin{enumerate}[nosep]
  \item Make the rubric stricter about coverage of essential context, not just the headline fact.
  \item Require the evaluator to check for \emph{every} key actor, \emph{every} motivation, and \emph{every} background event.
  \item Apply quantitative penalties: $-1$ per missing key actor, $-0.5$ per missing motivation or background event.
\end{enumerate}
\end{tcolorbox}

\noindent The resulting prompt decomposed relevance into exhaustive sub-criteria (actors, motivations, background events, factual propositions, redundancy) with a rigid penalty system. The regenerated binary questions reflected this over-specificity:

\begin{tcolorbox}[colback=white, colframe=red!30, fontupper=\small]
\textbf{Regenerated questions (relevance, failed iteration):}
\begin{enumerate}[nosep,leftmargin=*]
  \item Does the summary include \textit{every key actor} mentioned in the source?
  \item Does the summary include \textit{every motivation} for actions stated in the source?
  \item Does the summary include \textit{all background events} directly relevant to the headline?
  \item Does the summary contain \textit{every other factual proposition} (dates, locations, amounts)?
  \item Does the summary avoid irrelevant or redundant information?
\end{enumerate}
\end{tcolorbox}

\noindent \textbf{Why it fails:} Human annotators use a \emph{holistic} judgment for relevance---``did the summary capture the gist?''---with soft tolerance for missing minor details. The updated questions demand \emph{exhaustive} coverage, causing the model to rate almost all summaries as deficient. The resulting scores are systematically lower than human scores, destroying rank correlation. This illustrates a fundamental limitation: when the human evaluation criterion is inherently holistic and tolerant, decomposing it into strict atomic checks produces a harsher evaluator that diverges from human behavior.

\subsubsection{Example 4: IFBench --- Promptable vs.\ Computational Constraints}
\label{sec:example-ifbench}

\textbf{Result:} Format accuracy improved from 52\% to 69\%; count accuracy degraded from 63\% to 31\%.

The IFBench meta prompt starts minimal (22 characters: \texttt{"Respond to the query."}). As shown in \cref{tab:ifbench-bloat}, after 4 iterations of lesson extraction and prompt rewriting, it grows to 6,248 characters. The lessons fall into two categories:

\paragraph{Promptable lessons (effective).} For format and sentence constraints, lessons identify missing guidance that the model can follow:

\begin{tcolorbox}[colback=green!3, colframe=cyan!60, title={IFBench: Effective Lessons (Format/Sentence)}, fonttitle=\small\bfseries]
\small
\begin{itemize}[nosep]
  \item ``Output must be plain text with no markup unless explicitly required.''
  \item ``For repeat-type tasks, output ONLY the exact original request with the specified minimal change. Do not add explanations.''
  \item ``Obey the requested format exactly. Every line must follow the structure (indentation, list marker, newline) as described.''
\end{itemize}
\end{tcolorbox}

\paragraph{Computational lessons (ineffective).} For count and ratio constraints, lessons correctly diagnose the problem but prescribe unactionable instructions:

\begin{tcolorbox}[colback=red!3, colframe=red!40, title=IFBench: Ineffective Lessons (Count/Ratio), fonttitle=\small\bfseries]
\small
\begin{itemize}[nosep]
  \item ``Maintain a running counter for each required element and stop when the target count is reached.'' \hfill $\leftarrow$ \textit{model cannot execute}
  \item ``Programmatically verify the position of each token; if the count is wrong, rewrite until satisfied.'' \hfill $\leftarrow$ \textit{requires self-verification loop}
  \item ``Construct a structural outline and reference it when placing required words.'' \hfill $\leftarrow$ \textit{implicit reasoning, not enforced}
\end{itemize}
\end{tcolorbox}

\noindent The accumulation of these unactionable instructions causes \emph{prompt bloat}:

\begin{table}[h]
\centering
\small
\caption{IFBench meta prompt growth and its effect on accuracy.}
\label{tab:ifbench-bloat}
\begin{tabular}{l rr rr}
\toprule
& \textbf{Prompt} & & \multicolumn{2}{c}{\textbf{Accuracy (\%)}} \\
\cmidrule(lr){4-5}
\textbf{Iteration} & \textbf{Size} & \textbf{Lessons} & Format & Count \\
\midrule
0 (baseline) & 22 chars  & --- & 52 & 63 \\
1            & 1,841     & 10  & 57 & 63 \\
2            & 3,425     & 10  & 52 & 52 \\
3            & 4,890     & 10  & \textbf{69} & 42 \\
4            & 6,248     & 10  & 48 & 31 \\
\bottomrule
\end{tabular}
\vspace{10pt}
\end{table}

\noindent \textbf{Insight:} At iteration 3, the prompt is large enough to contain useful format guidance but not yet so bloated that attention competition degrades all categories. By iteration 4, the accumulated computational instructions (which the model cannot follow) create noise that interferes with previously-working format guidance, causing a collapse across all categories. This reveals a \emph{carrying capacity} for prompt-based optimization: beyond a critical prompt length, additional instructions become counterproductive regardless of their correctness.

\section{Experimental Setups}
\label{app:exp_setup}

\rwparagraph{SummEval --- Evaluator Prompt Optimization}
We optimize the evaluator prompt for gpt-oss-120b on all four SummEval dimensions: coherence, consistency, fluency, and relevance. SummEval contains 1,600 items (100 documents $\times$ 16 summarization systems) with human Likert ratings on a 1--5 scale.
\begin{itemize}[nosep,leftmargin=*]
  \item \textbf{Data split.} We randomly sample 10 items per system (seed $= 42$), yielding 160 development items for lesson extraction and 1,440 test items for evaluation. The development set spans 82 of the 100 documents, providing broad coverage while keeping the update loop manageable.
  \item \textbf{Models.} The target evaluator is gpt-oss-120b with temperature $0$. For self-update, the same model also serves as the note-taker for lesson extraction, semantic deduplication, and prompt rewriting. For cross-model update, Claude Sonnet 4 serves as both the source evaluator and the note-taker, again with temperature $0$.
  \item \textbf{Procedure.} Each iteration follows Algorithm~\ref{fig:algorithm}: (1) evaluate the development and test sets with the current prompt and binary questions; (2) for self-update, identify items where the model score diverges most from the human score ($|s_{\mathrm{model}} - s_{\mathrm{human}}| > 0.3$ after normalization), while for cross-model update we identify question-level disagreements between the source and target evaluators; (3) extract lessons from these failures or disagreements in batches, semantically deduplicate them with an LLM, and retain up to 10 unique lessons; (4) rewrite the evaluator prompt in a single LLM call incorporating all retained lessons; and (5) regenerate binary questions from the updated prompt for self-update.
  \item \textbf{Early stopping.} We run up to 5 iterations and stop when the test-set Spearman $\rho$ decreases relative to the previous iteration.
  \item \textbf{Metric.} We report pooled Spearman rank correlation across all test items rather than a per-document average, since per-document averaging discards documents with fewer than two systems in the sparse development split.
\end{itemize}

\rwparagraph{IFBench --- Generation Prompt Optimization}
We optimize the generation meta-prompt for gpt-oss-120b on IFBench, an instruction-following benchmark with 290 test cases spanning 56 constraint types across 7 categories: count, words, format, ratio, sentence, repeat, and custom. Each case includes a programmatic verification function.
\begin{itemize}[nosep,leftmargin=*]
  \item \textbf{Data split.} The development set contains 56 samples, one per constraint type, preferring previously failed cases; the test set contains the remaining 238 samples.
  \item \textbf{Models.} The generator is gpt-oss-120b with temperature $0$. For self-update, the judge is also gpt-oss-120b. For cross-model update, the judge is Claude Sonnet 4.
  \item \textbf{Binary question decomposition.} For each development sample, we convert the IFBench constraint specification into a natural-language description and decompose it into binary yes/no questions using the \bindeval meta-prompt. For example, a constraint requiring one occurrence of \texttt{door} and two occurrences of \texttt{bread} becomes questions such as whether the response includes \texttt{door} exactly once and \texttt{bread} exactly twice.
  \item \textbf{Procedure.} Each iteration: (1) generate responses on all 290 samples with the current meta-prompt; (2) evaluate the development responses with an LLM judge using binary questions; (3) extract and deduplicate lessons from development failures; (4) rewrite the generation meta-prompt; and (5) evaluate on the test set using the official IFBench verification functions in strict mode.
  \item \textbf{Iterations.} Self-update runs for 5 iterations. Cross-model update stops after 2 iterations because test accuracy decreases.
\end{itemize}

\newpage
\section{Automatic Prompt Update Algorithm}
\begin{algorithm}[!ht]
\caption{Iterative Prompt Update via Binary Question Disagreement}
\label{fig:algorithm}
\DontPrintSemicolon
\SetKwInput{KwIn}{Input}
\SetKwInput{KwOut}{Output}
\KwIn{Source evaluator $E_{\mathrm{src}}$, target evaluator $E_{\mathrm{tgt}}$ with initial prompt $P_E^{(0)}$, binary questions $Q = \{q_1, \ldots, q_N\}$, test data $\{(x_j, y_j)\}_{j=1}^{J}$, note-taker LLM $L_{\mathrm{note}}$, updater LLM $L_{\mathrm{update}}$, tolerance $\epsilon$, max iterations $T$}
\KwOut{Updated prompt $P_E^{(T)}$}
\For{$t \leftarrow 1$ \KwTo $T$}{
    \tcp{Step 1: Evaluate with both models}
    \ForEach{$(x_j, y_j)$ in test data}{
        $A_j^{\mathrm{src}} \leftarrow \{f_{E_{\mathrm{src}}}(x_j, y_j, q_i)\}_{i=1}^{N}$\;
        $A_j^{\mathrm{tgt}} \leftarrow \{f_{E_{\mathrm{tgt}}}(x_j, y_j, q_i; P_E^{(t-1)})\}_{i=1}^{N}$\;
    }
    \tcp{Check convergence}
    \ForEach{dimension $d$ in $D$}{
        $S_d^{\mathrm{tgt}} \leftarrow (1/|Q_d|) * \sum_{q_i \in Q_d} \mathrm{mean}_j[A_j^{\mathrm{tgt}}(q_i)]$\;
        $S_d^{\mathrm{src}} \leftarrow (1/|Q_d|) * \sum_{q_i \in Q_d} \mathrm{mean}_j[A_j^{\mathrm{src}}(q_i)]$\;
    }
    \If{$|S_d^{\mathrm{tgt}} - S_d^{\mathrm{src}}| < \epsilon$ for all $d$}{
        \Return{$P_E^{(t-1)}$}\tcp*{Converged}
    }
    \tcp{Step 2: Identify disagreements}
    \ForEach{$(x_j, y_j)$ in test data}{
        $\Delta_j \leftarrow \{q_i : A_j^{\mathrm{src}}(q_i) \neq A_j^{\mathrm{tgt}}(q_i)\}$\;
    }
    \tcp{Step 3: Extract lessons from disagreements}
    $L_{\mathrm{all}} \leftarrow$ empty list\;
    \ForEach{$j$ where $|\Delta_j| > 0$}{
        $L_j \leftarrow L_{\mathrm{note}}(x_j, y_j, A_j^{\mathrm{src}}, A_j^{\mathrm{tgt}}, \Delta_j)$\;
        $L_{\mathrm{all}} \leftarrow L_{\mathrm{all}} + L_j$\;
    }
    \tcp{Step 4: Semantic deduplication}
    $M \leftarrow$ empty list\tcp*{Lesson memory}
    \ForEach{$l_{\mathrm{new}}$ in $L_{\mathrm{all}}$}{
        $(\mathrm{is\_dup}, \mathrm{merge\_idx}, \mathrm{merged}) \leftarrow \mathrm{Dedup\_LLM}(l_{\mathrm{new}}, M)$\;
        \eIf{is\_dup}{
            $M[\mathrm{merge\_idx}] \leftarrow \mathrm{merged}$\;
        }{
            $M.\mathrm{append}(l_{\mathrm{new}})$\;
        }
    }
    $L_{\mathrm{unique}} \leftarrow M$\;
    \tcp{Step 5: Update target evaluator prompt}
    $P_E^{(t)} \leftarrow P_E^{(t-1)}$\;
    \ForEach{$l_k$ in $L_{\mathrm{unique}}$}{
        $(s_k, s_k') \leftarrow L_{\mathrm{update}}(P_E^{(t)}, l_k)$\;
        $P_E^{(t)} \leftarrow P_E^{(t)}.\mathrm{replace}(s_k, s_k')$\;
    }
}
\Return{$P_E^{(T)}$}
\end{algorithm}

\section{Results}

\subsection{\bindeval Evaluation Results on Topical-Chat}
\label{app:topical_chat}

Table~\ref{tab:topicalchat} establishes four main findings. First, \bindeval (Claude) is the strongest overall method on Topical-Chat, with the best average Spearman and Kendall correlations (0.632 / 0.525), outperforming G-Eval (GPT-4), UniEval (T5), and all lexical baselines. This indicates that multi-question binary decomposition is a strong evaluation paradigm for dialogue, where quality depends on several partially independent criteria rather than a single aggregate impression.

Second, different methods are strongest on different dimensions. \bindeval (Claude) performs best on the most subjective dimensions, naturalness and engagingness, improving over G-Eval (GPT-4) by 0.137 and 0.113 in Spearman correlation. By contrast, UniEval (T5) and G-Eval remain stronger on coherence, where a more holistic representation may better capture global logical flow. Groundedness is comparatively method-agnostic: all LLM-based evaluators are within a narrow range, suggesting that this dimension is easier to capture regardless of evaluation format.

Third, evaluator quality is a first-order factor. \bindeval (gpt-oss) reaches 0.539 / 0.450 on average, close to G-Eval (gpt-oss) at 0.541 / 0.478 and far above UniEval (gpt-oss) at 0.144 / 0.132, but still well below the Claude-based version. In other words, good question decomposition helps, but the evaluator must still be capable of answering conversational questions with enough nuance. This is especially clear for UniEval (gpt-oss): a single binary question often collapses to nearly constant outputs, such as naturalness at 0.

Fourth, question design is helpful but bounded. The \bindeval (gpt-oss) remains competitive because multiple binary questions create useful score granularity even when single-score calibration is weak, but it still does not match the Claude-based evaluator. Overall, the Topical-Chat results show that decomposition is particularly valuable for subjective dialogue qualities, while still depending on evaluator strength for best performance.

\begin{figure}[!h]
\centering
\includegraphics[width=0.95\textwidth]{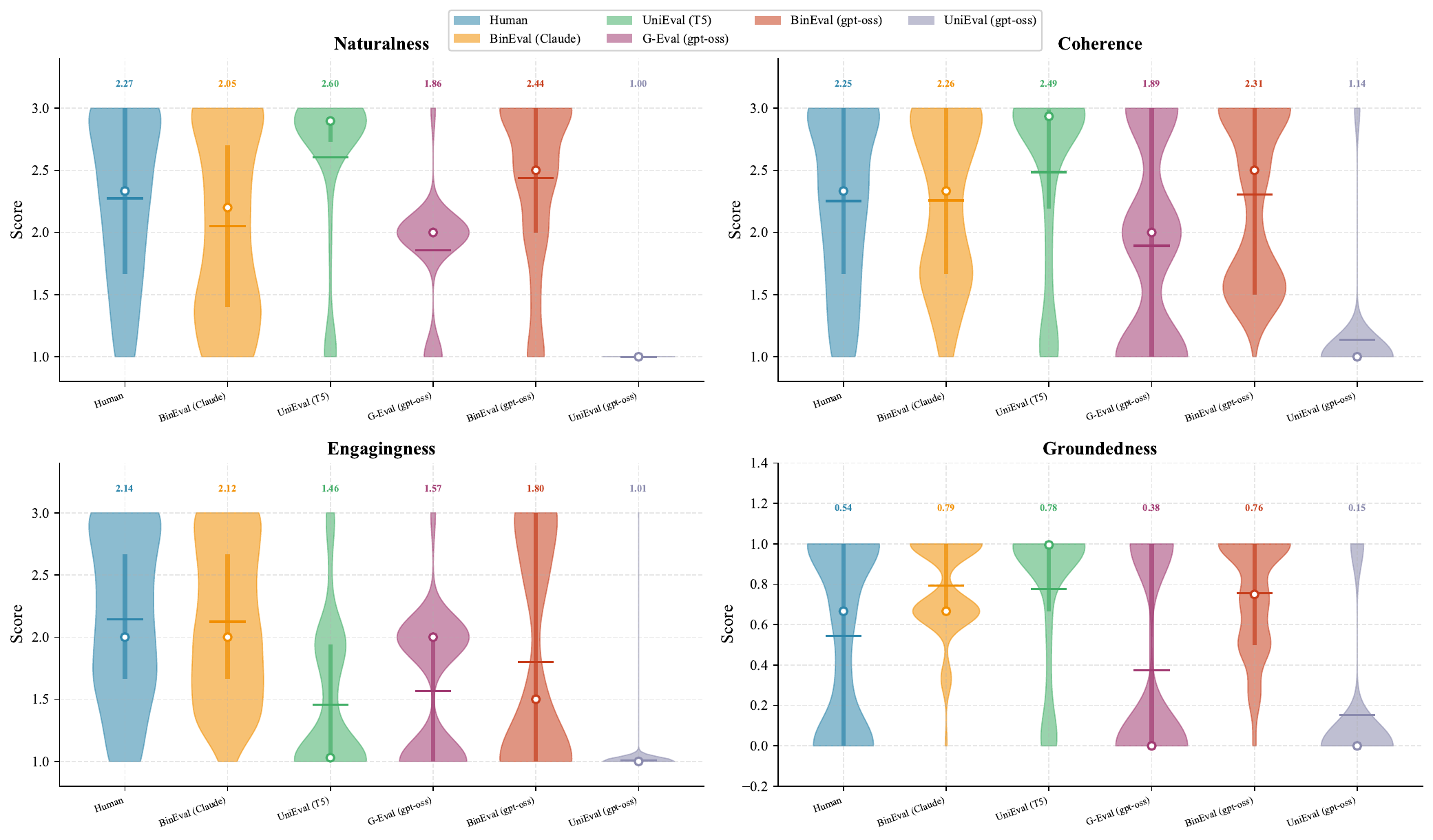}
\caption{Per-dimension score distributions on Topical-Chat. \bindeval (Claude) most closely tracks the human distributions across naturalness, coherence, engagingness, and groundedness. UniEval (T5) exhibits clear ceiling effects, especially outside engagingness; \bindeval (gpt-oss) remains more discriminative than the other gpt-oss-based baselines; and UniEval (gpt-oss) is nearly flat across dimensions.}
\label{fig:topical-violin-dim}
\end{figure}

The violin plots reinforce these trends. In Figure~\ref{fig:topical-violin-dim}, \bindeval (Claude) most closely matches the human spread and skew across all four dimensions, preserving both the broader dispersion on engagingness and the more concentrated but still non-degenerate distributions on naturalness, coherence, and groundedness. UniEval (T5) remains high but noticeably compressed, with a pronounced ceiling effect on naturalness, coherence, and groundedness and much weaker alignment on engagingness. Among the gpt-oss-based evaluators, \bindeval (gpt-oss) is more conservative than the Claude version but still retains meaningful variation across examples, whereas G-Eval (gpt-oss) is more compressed and UniEval (gpt-oss) is nearly degenerate across all dimensions, providing very little discrimination.

\begin{table}[!t]
\centering
\small
\caption{Turn-level Spearman $\rho$ / Kendall $\tau$ correlations on Topical-Chat.}
\label{tab:topicalchat}
\begin{tabular}{lccccccc}
\toprule
\textbf{Method} & \textbf{Naturalness} & \textbf{Coherence} & \textbf{Engagingness} & \textbf{Groundedness} & \textbf{Average} \\
\midrule
ROUGE-L & 0.176 / 0.146 & 0.193 / 0.203 & 0.295 / 0.300 & 0.310 / 0.327 & 0.243 / 0.244 \\
BLEU-4 & 0.180 / 0.175 & 0.131 / 0.235 & 0.232 / 0.316 & 0.213 / 0.310 & 0.189 / 0.259 \\
METEOR & 0.212 / 0.191 & 0.250 / 0.302 & 0.367 / 0.439 & 0.333 / 0.391 & 0.290 / 0.331 \\
BERTScore & 0.226 / 0.209 & 0.214 / 0.233 & 0.317 / 0.335 & 0.291 / 0.317 & 0.262 / 0.273 \\
UniEval (T5) & 0.455 / 0.330 & \textbf{0.602} / 0.455 & 0.573 / 0.430 & 0.577 / 0.453 & 0.552 / 0.417 \\
G-Eval (GPT-4) & 0.549 / \textbf{0.565} & 0.594 / \textbf{0.605} & 0.627 / \textbf{0.631} & 0.531 / \textbf{0.551} & 0.575 / \textbf{0.588} \\
{G-Eval (gpt-oss)} & 0.422 / 0.368 & 0.526 / 0.444 & 0.642 / 0.560 & 0.576 / 0.539 & 0.541 / 0.478 \\
UniEval (gpt-oss) & 0.000 / 0.000 & 0.132 / 0.116 & 0.073 / 0.064 & 0.370 / 0.346 & 0.144 / 0.132 \\
\textbf{\bindeval (gpt-oss)} & 0.483 / 0.420 & 0.447 / 0.359 & 0.648 / 0.532 & \textbf{0.578} / 0.488 & 0.539 / 0.450 \\
\textbf{\bindeval (Claude)} & \textbf{0.686 / 0.565} & 0.564 / 0.447 & \textbf{0.740} / 0.606 & 0.538 / 0.485 & \textbf{0.632} / 0.525 \\
\bottomrule
\end{tabular}
\vspace{10pt}
\end{table}

\begin{figure}[!h]
\centering
\includegraphics[width=0.95\textwidth]{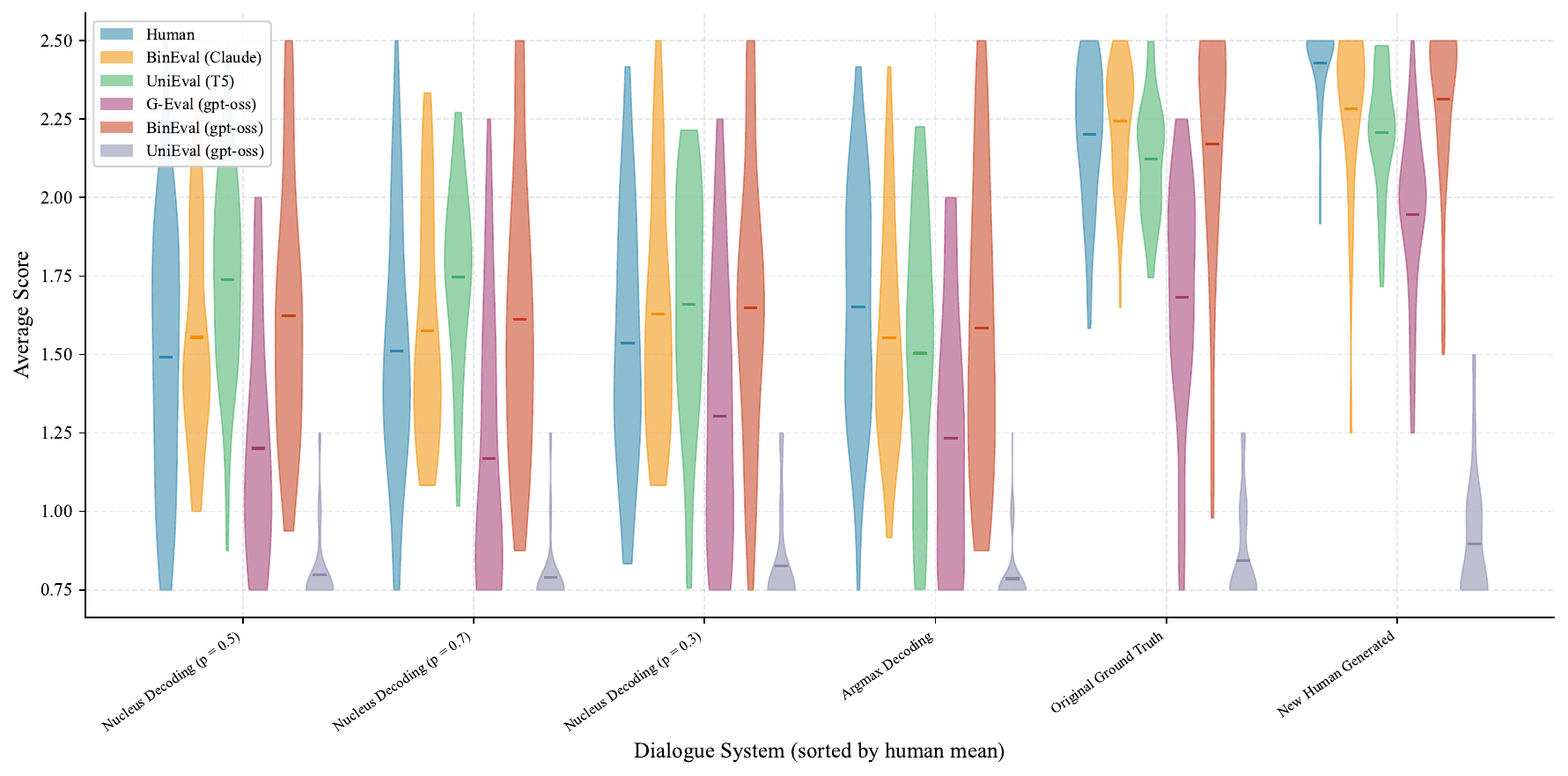}
\caption{Per-system score distributions on Topical-Chat. \bindeval (Claude) best preserves the human ordering of systems while maintaining realistic within-system spread. \bindeval (gpt-oss) follows the broad ranking but is more conservative in absolute score level, G-Eval (gpt-oss) compresses low- and mid-performing systems, and UniEval (gpt-oss) is nearly uninformative because its scores are almost constant across systems.}
\label{fig:topical-violin-system}
\vspace{10pt}
\end{figure}

Figure~\ref{fig:topical-violin-system} shows the same pattern at the system level. \bindeval (Claude) best preserves the human ordering of systems, separating stronger systems from weaker ones while keeping realistic within-system variation rather than collapsing all outputs into a narrow high-scoring band. \bindeval (gpt-oss) also tracks the broad ranking but with lower absolute scores, suggesting that decomposition still helps even when the underlying evaluator is weaker. By contrast, G-Eval (gpt-oss) compresses much of the low-to-mid range, and UniEval (gpt-oss) is nearly flat across systems. Together, these plots illustrate the central advantage of decomposition: multiple targeted questions produce more realistic and discriminative score variation than a single holistic or near-Boolean judgment.

\subsection{\bindeval Evaluation Results on QAGS}
\label{app:qags}

\begin{table}[!t]
\centering
\small
\caption{Correlation results on QAGS. Pearson $r$ / Spearman $\rho$ / Kendall $\tau$ for QAGS-CNN, QAGS-XSUM, and their average.}
\label{tab:qags}
\begin{tabular}{lccccccccc}
\toprule
\textbf{Metrics} & \multicolumn{3}{c}{\textbf{QAGS-CNN}} & \multicolumn{3}{c}{\textbf{QAGS-XSUM}} & \multicolumn{3}{c}{\textbf{Average}} \\
\cmidrule(lr){2-4} \cmidrule(lr){5-7} \cmidrule(lr){8-10}
 & $r$ & $\rho$ & $\tau$ & $r$ & $\rho$ & $\tau$ & $r$ & $\rho$ & $\tau$ \\
\midrule
ROUGE-L & 0.357 & 0.324 & 0.254 & 0.024 & -0.011 & -0.009 & 0.190 & 0.156 & 0.122 \\
BERTScore & 0.576 & 0.505 & 0.399 & 0.024 & 0.008 & 0.006 & 0.300 & 0.256 & 0.202 \\
MoverScore & 0.414 & 0.347 & 0.271 & 0.054 & 0.044 & 0.036 & 0.234 & 0.195 & 0.153 \\
FactCC & 0.416 & 0.484 & 0.376 & 0.297 & 0.259 & 0.212 & 0.356 & 0.371 & 0.294 \\
QAGS & 0.545 & \dashscore & \dashscore & 0.175 & \dashscore & \dashscore & 0.375 & \dashscore & \dashscore \\
BARTScore & \textbf{0.735} & 0.680 & 0.557 & 0.184 & 0.159 & 0.130 & 0.459 & 0.420 & 0.343 \\
CTC & 0.619 & 0.564 & 0.450 & 0.309 & 0.295 & 0.242 & 0.464 & 0.430 & 0.346 \\
UniEval (T5) & 0.682 & 0.662 & 0.532 & 0.461 & 0.488 & 0.399 & 0.571 & 0.575 & 0.465 \\
G-Eval (GPT-4) & 0.631 & {0.685} & {0.591} & \textbf{0.558} & {0.537} & \textbf{0.472} & {0.599} & {0.611} & {0.525} \\
G-Eval (gpt-oss) & 0.045 & 0.028 & 0.027 & 0.236 & 0.236 & 0.236 & 0.140 & 0.132 & 0.131 \\
UniEval (gpt-oss) & 0.415 & 0.382 & 0.357 & 0.490 & 0.490 & 0.490 & 0.452 & 0.436 & 0.424 \\
\textbf{\bindeval (gpt-oss)} & 0.651 & 0.642 & 0.551 & 0.435 & 0.483 & 0.433 & 0.543 & 0.563 & 0.492 \\
\textbf{\bindeval (Claude)} & 0.665 & \textbf{0.702} & \textbf{0.597} & 0.543 & \textbf{0.539} & 0.470 & \textbf{0.604} & \textbf{0.620} & \textbf{0.534} \\
\bottomrule
\end{tabular}
\vspace{10pt}
\end{table}

Table~\ref{tab:qags} highlights the setting where question decomposition helps most. \bindeval (Claude) is the strongest method overall, with the best average Pearson, Spearman, and Kendall correlations (0.604 / 0.620 / 0.534). It is strongest on rank-based metrics for both datasets, achieving the top Spearman values on CNN/DM and XSum (0.702 and 0.539), while remaining competitive in Pearson correlation against stronger regression-style baselines such as BARTScore on CNN/DM and G-Eval (GPT-4) on XSum. \bindeval (gpt-oss) is also robust, reaching 0.543 / 0.563 / 0.492 on average and substantially outperforming the other gpt-oss based evaluators.

The dataset-level breakdown is also informative. CNN/DM is the easier split: most strong evaluators achieve reasonably high correlations, and both \bindeval variants perform well there, with \bindeval (Claude) at 0.665 / 0.702 / 0.597 and \bindeval (gpt-oss) at 0.651 / 0.642 / 0.551. XSum is harder for every method, but the relative pattern remains the same: \bindeval (Claude) still gives the best Spearman correlation (0.539), narrowly ahead of G-Eval (GPT-4) at 0.537, while \bindeval (gpt-oss) remains competitive at 0.483. The additional gpt-oss baselines make the advantage of decomposition especially clear. G-Eval (gpt-oss) nearly collapses on QAGS, reaching only 0.140 / 0.132 / 0.131 on average, and UniEval (gpt-oss) recovers some signal because factual consistency is closer to a binary property, but still trails both \bindeval variants. In short, QAGS shows that decomposition is most valuable when a single holistic prompt fails to preserve enough ranking granularity.

\begin{figure}[!h]
\centering
\includegraphics[width=0.95\textwidth]{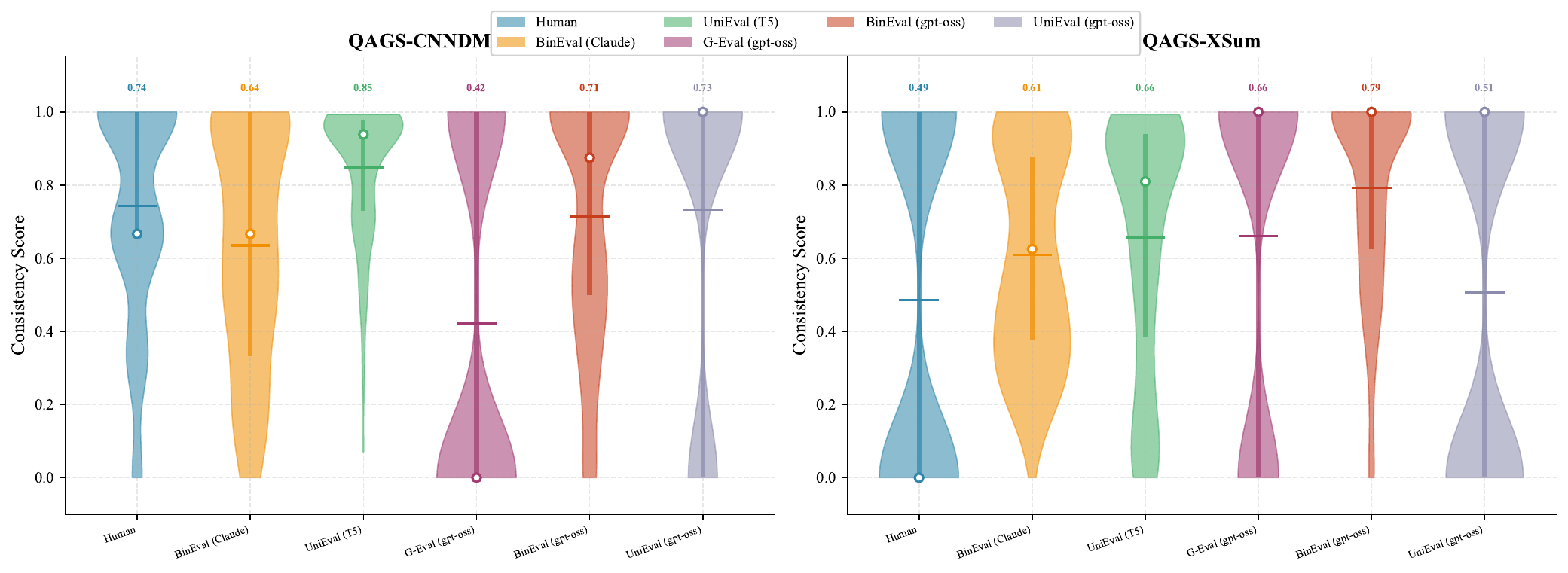}
\caption{Per-dataset score distributions on QAGS for human ratings, \bindeval, and UniEval.}
\label{fig:qags-violin-system}
\end{figure}

Figure~\ref{fig:qags-violin-system} supports the same conclusion in distributional form. For both CNN/DM and XSum, the human ratings are distinctly bimodal, with substantial mass near both 0 and 1. \bindeval (Claude) is the method that most clearly preserves this structure: it keeps broad support across the full range instead of collapsing toward the top of the scale. \bindeval (gpt-oss) is somewhat more conservative but still retains visible spread and separation. By contrast, UniEval (T5) is strongly overconfident on both datasets, with most of its mass concentrated near high scores, while G-Eval (gpt-oss) and UniEval (gpt-oss) become almost binary in the wrong way---they place much of the distribution at the extremes with very limited intermediate variation. This matters because a useful factual evaluator must distinguish mildly flawed summaries from clearly inconsistent ones, not just separate obviously correct cases from obviously incorrect ones.

\begin{figure}[!h]
\centering
\includegraphics[width=0.95\textwidth]{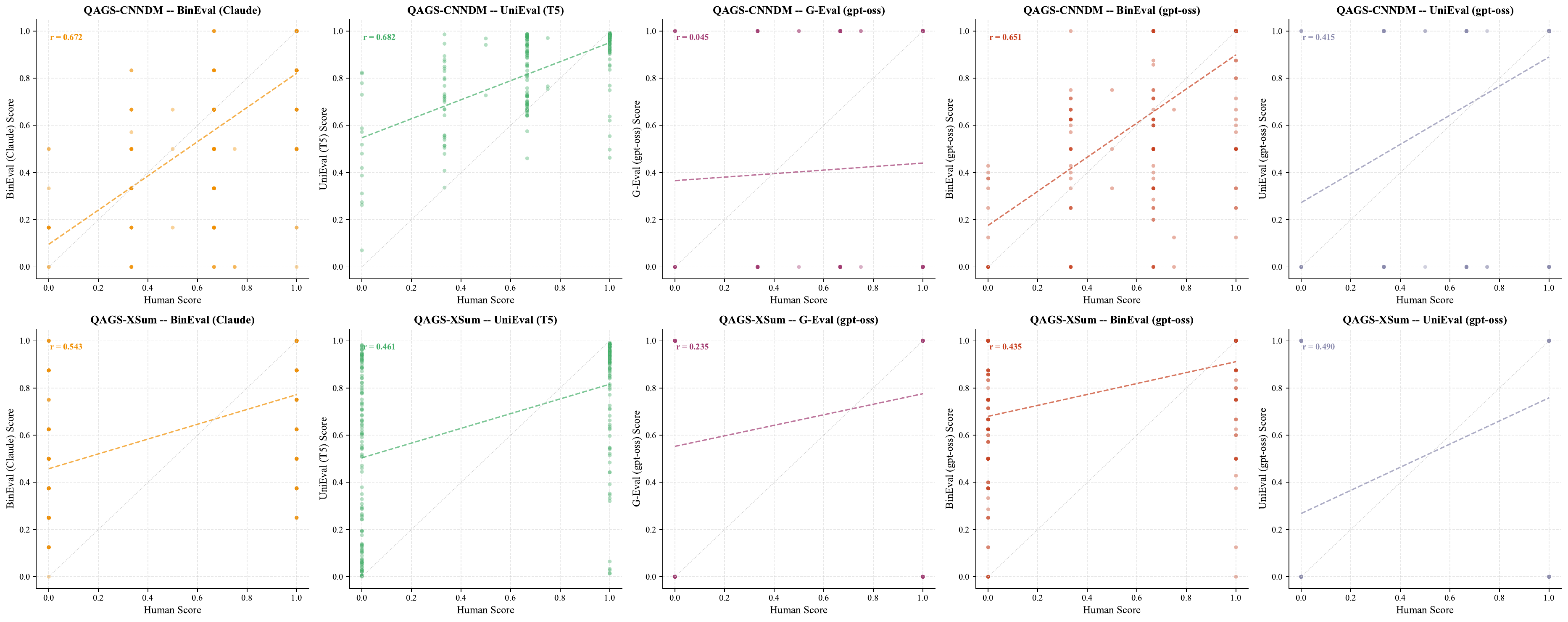}
\caption{Per-summary scatter plots against human consistency scores on QAGS.}
\label{fig:qags-scatter-system}
\end{figure}

Figure~\ref{fig:qags-scatter-system} makes the ranking behavior even clearer. \bindeval (Claude) shows the cleanest positive trend on both datasets, with fitted lines that track the diagonal substantially better than the other methods. \bindeval (gpt-oss) follows the same pattern, though with more dispersion, matching its strong but slightly lower correlations in Table~\ref{tab:qags}. UniEval (T5) produces a positive trend but compresses many predictions into a narrow upper band, which limits discrimination despite decent correlation. G-Eval (gpt-oss) is nearly flat on CNN/DM and only weakly increasing on XSum, while UniEval (gpt-oss) exhibits only coarse, quantized outputs. Together, the table and figures show that the key benefit of decomposition on QAGS is not only better correlation, but also better use of the score range: \bindeval assigns meaningfully different scores to different kinds of factual errors instead of collapsing them into a small set of near-identical predictions.

\section{Binary Questions for SummEval}
\label{app:binary_questions}

Tables~\ref{tab:questions-coherence}--\ref{tab:questions-relevance} list the binary questions auto-generated by \bindeval for each SummEval evaluation dimension. These are the questions referenced in \cref{sec:analysis} and \cref{fig:correlation}. Each question is designed so that ``yes'' indicates the output satisfies the criterion and ``no'' indicates a violation.

\begin{table}[H]
\centering
\small
\caption{Binary questions for \textbf{Coherence} on SummEval (8 questions).}
\label{tab:questions-coherence}
\begin{tabular}{cp{0.78\columnwidth}c}
\toprule
\textbf{ID} & \textbf{Question} & \textbf{Yes-rate} \\
\midrule
Q1 & Does the summary have a well-defined structure (e.g., a clear beginning, middle, and/or end) rather than appearing randomly assembled? & 0.66 \\
Q2 & Are the sentences in the summary arranged in a sensible and logical order? & 0.68 \\
Q3 & Does the summary avoid being a mere heap of loosely related facts or information? & 0.43 \\
Q4 & Do the sentences in the summary flow logically from one to the next, with clear transitions or connections between them? & 0.44 \\
Q5 & Does the summary maintain a unified focus on a single main topic rather than drifting across multiple unrelated subjects? & 0.96 \\
Q6 & Does the summary cover the main topic of the news article? & 0.85 \\
Q7 & Does the summary cover the key points of the news article? & 0.10 \\
Q8 & Does the summary present information in a clear manner that is easy to follow and understand? & 0.61 \\
\bottomrule
\end{tabular}
\end{table}

\begin{table}[H]
\centering
\small
\caption{Binary questions for \textbf{Consistency} on SummEval (7 questions).}
\label{tab:questions-consistency}
\begin{tabular}{cp{0.78\columnwidth}c}
\toprule
\textbf{ID} & \textbf{Question} & \textbf{Yes-rate} \\
\midrule
Q1 & Are all statements in the summary entailed by or supported by the source article? & 0.75 \\
Q2 & Is the summary free of factual errors when compared to the source article? & 0.82 \\
Q3 & Is the summary free of hallucinated facts (i.e., information that is fabricated and not present in the source article)? & 0.81 \\
Q4 & Are all named entities (people, organizations, locations) in the summary accurately represented as they appear in the source article? & 0.91 \\
Q5 & Are all numerical claims (dates, statistics, quantities, amounts) in the summary consistent with the source article? & 0.95 \\
Q6 & Are the causal relationships and event sequences described in the summary consistent with those in the source article? & 0.87 \\
Q7 & Does the summary avoid misrepresenting or distorting the meaning of information from the source article? & 0.76 \\
\bottomrule
\end{tabular}
\end{table}

\begin{table}[H]
\centering
\small
\caption{Binary questions for \textbf{Fluency} on SummEval (7 questions).}
\label{tab:questions-fluency}
\begin{tabular}{cp{0.78\columnwidth}c}
\toprule
\textbf{ID} & \textbf{Question} & \textbf{Yes-rate} \\
\midrule
Q1 & Is the summary free of grammatical errors? & 0.54 \\
Q2 & Is the summary free of spelling errors? & 0.71 \\
Q3 & Is the summary free of punctuation errors? & 0.33 \\
Q4 & Does the summary use appropriate and natural word choices? & 0.81 \\
Q5 & Does the summary have well-structured sentences that are easy to follow? & 0.70 \\
Q6 & Does the summary read smoothly and sound natural overall? & 0.52 \\
Q7 & Is the summary easy to understand without requiring re-reading due to language issues? & 0.76 \\
\bottomrule
\end{tabular}
\end{table}

\begin{table}[H]
\centering
\small
\caption{Binary questions for \textbf{Relevance} on SummEval (5 questions).}
\label{tab:questions-relevance}
\begin{tabular}{cp{0.78\columnwidth}c}
\toprule
\textbf{ID} & \textbf{Question} & \textbf{Yes-rate} \\
\midrule
Q1 & Does the summary address the main topic or central event of the source article? & 0.95 \\
Q2 & Does the summary cover at least some of the key points or important details of the source article? & 0.50 \\
Q3 & Is the summary free from significant redundancy, such as repeating the same point multiple times in different words? & 0.64 \\
Q4 & Is the summary free from excessive trivial or unimportant details that dilute the coverage of main points? & 0.72 \\
Q5 & Does the summary prioritize the most newsworthy or significant information from the source rather than focusing on minor or tangential aspects? & 0.49 \\
\bottomrule
\end{tabular}
\end{table}

\end{document}